\documentclass[10pt,journal,compsoc]{IEEEtran}

\usepackage{color}
\usepackage{caption}
\usepackage{times}
\usepackage{helvet}
\usepackage{courier}
\usepackage{amsmath}
\usepackage{amssymb}
\usepackage{amsthm}
\usepackage{graphicx}
\usepackage{array}
\usepackage{booktabs}
\usepackage{amsopn}
\usepackage{amsmath,bm}
\usepackage{algorithm}
\usepackage{algorithmic}
\usepackage{enumerate}
\usepackage{color}
\usepackage{multirow}
\usepackage{bbding}

\newcommand{\eg}{\emph{e.g.}}
\newcommand{\etal}{\emph{et al.}}
\newcommand{\ie}{\emph{i.e.}}

\newcommand{\wrt}{w.r.t. }

\newcommand{\vect}{\text{vec}}

\graphicspath{{figs/}}
\begin{document}

\title{Learning Versatile Convolution Filters for Efficient Visual Recognition}
\author{Kai Han, Yunhe Wang,~\IEEEmembership{Member,~IEEE}, Chang Xu,~\IEEEmembership{Member,~IEEE}, Chunjing Xu, Enhua Wu,\\and Dacheng Tao,~\IEEEmembership{Fellow,~IEEE}
\IEEEcompsocitemizethanks{\IEEEcompsocthanksitem 
Kai Han and Enhua Wu are with State Key Laboratory of Computer Science, Institute of Software, Chinese Academy of Sciences, Beijing, 100190, China. They are also with University of Chinese Academy of Sciences, Beijing, 100049, China. Kai Han is also with Noah's Ark Lab, Huawei Technologies, and Enhua Wu is also with FST, University of Macau. Email: hankai@ios.ac.cn, ehwu@umac.mo.
\IEEEcompsocthanksitem Yunhe Wang and Chunjing Xu are with Noah's Ark Lab, Huawei Technologies. E-mail: yunhe.wang@huawei.com, xuchunjing@huawei.com.
\IEEEcompsocthanksitem Chang Xu and Dacheng Tao are with School of Computer Science, in the Faculty of Engineering, at The University of Sydney, 6 Cleveland St, Darlington, NSW 2008, Australia. Email: c.xu@sydney.edu.au, dacheng.tao@sydney.edu.au.
\IEEEcompsocthanksitem Kai Han and Yunhe Wang contributed equally to this work.
\IEEEcompsocthanksitem Correspondence to Chang Xu and Enhua Wu.
}
}

\IEEEtitleabstractindextext{%
	\begin{abstract}
		This paper introduces versatile filters to construct efficient convolutional neural networks that are widely used in various visual recognition tasks. Considering the demands of efficient deep learning techniques running on cost-effective hardware, a number of methods have been developed to learn compact neural networks. Most of these works aim to slim down filters in different ways, \eg,~investigating small, sparse or quantized filters. In contrast, we treat filters from an additive perspective. A series of secondary filters can be derived from a primary filter with the help of binary masks. These secondary filters all inherit in the primary filter without occupying more storage, but once been unfolded in computation they could significantly enhance the capability of the filter by integrating information extracted from different receptive fields. Besides spatial versatile filters, we additionally investigate versatile filters from the channel perspective. Binary masks can be further customized for different primary filters under orthogonal constraints. We conduct theoretical analysis on network complexity and an efficient convolution scheme is introduced. Experimental results on benchmark datasets and neural networks demonstrate that our versatile filters are able to achieve comparable accuracy as that of original filters, but require less memory and computation cost.
	\end{abstract}
	
	\begin{IEEEkeywords}
		CNN compression, CNN speed-up, Versatile filters.
\end{IEEEkeywords}}

\maketitle

\IEEEdisplaynontitleabstractindextext

\IEEEpeerreviewmaketitle

\IEEEraisesectionheading{\section{Introduction}\label{Sec:Introduction}}
\IEEEPARstart{C}{onsiderable} computer vision applications (\eg,~image classification~\cite{VGGnet}, object detection~\cite{renNIPS15fasterrcnn}, and image segmentation~\cite{FCN}) have received remarkable progress with the help of convolutional neural networks (CNNs) in last decade. From the pioneering AlexNet~\cite{alexnet} to the recent ResNeXt~\cite{ResNeXt}, the storage of networks is slightly saved, but the classification accuracy has been continuously improved. This performance improvement comes from sophisticatedly designed calculations in these networks, \eg,~residual modules in ResNet~\cite{he2016deep} and inception modules in GoogleNet~\cite{GoogleNet}. These networks are widely used in the scenario of abundant computation and storage resources, but cannot be easily deployed on mobile platforms, such as smartphones and cameras. Taking ResNet-50~\cite{he2016deep} with 54 convolutional layers as an example, about $97$\emph{MB} memory is required to store all filters and about $4.1\times10^9$ times of floating number multiplications have to be operated for an image. 

Over the years, different techniques have been proposed to tackle the contradiction between resources supply of low performance devices and demands of heavy neural networks. Various methods have been developed to explore and eliminate redundancy in pre-trained CNNs. For example, Han~\etal~\cite{pruning} discarded subtle weights in convolution filters, Wang~\etal~\cite{CNNpackNIPS} investigated redundancy between weights, Figurnov~\etal~\cite{sparse1} removed redundant connections between input data and filters, Wang~\etal~\cite{BeyondFilters} explored compact feature maps for deep neural networks, and Wen~\etal~\cite{sparse2} investigated the sparsity of weights in CNNs from several aspects. There are also methods to approximate the original neural networks using compact data or network structures, \eg,~quantization and binarization~\cite{3binary,binary}, matrix decomposition~\cite{SVD}, and teacher student learning paradigm~\cite{Distill}. Instead of patching pre-trained CNNs, some highly efficient network architectures have been designed for applications on mobile devices. For example, ResNeXt~\cite{ResNeXt} aggregated a set of transformations with the same topology, Xception~\cite{xception} and MobileNet~\cite{mobilenet,mobilev2} used separable convolutions with $1\times 1$ filters, and ShuffleNet~\cite{shufflenet,shufflev2} encouraged pointwise group convolutions and channel shuffle operations.

Most of these existing works learn efficient CNNs through slimming down filters, \eg,~making great use of smaller filters (\eg,~$1\times 1$ filters) and developing various (\eg,~sparse and low-rank) approximation of filters. Given such lightweight filters, the network performance is struggling to keep up, due to limited capacity of $1\times 1$ filters or approximation error of filters. Rather than subtracting (\ie,~slimming down filters), another thing to consider is adding. We must ask whether the value of a normal filter has already been maximally explored and can a normal filter take more roles than usual.

In this paper, we propose versatile filters for efficient convolutional neural networks. We sample a series of smaller secondary filters from a primary filter according to some pre-defined rules. These secondary filters inherit weights from the primary filter, but they will have different receptive fields and extract features from the spatial dimension. The neural network is composed of primary filters, while the strength of the network will be fully disclosed through secondary filters in computation. Specifically, we develop versatile filters in both spatial and channel dimensions. Binary masks play an important role to unfold primary filters for distinct secondary filters. Beyond hand crafted binary masks to produce versatile filters, we further learn to be versatile via learnable binary masks. Orthogonal constraints are exploited to promote the diversity of these secondary filters. Both primary filers and binary masks are learned in an end-to-end manner through a specifically designed optimization method. The resulting neural network is lightweight, as only a few primary filters are in floating-point resolution while associated masks are in 1-bit format. Memory cost is discussed in detail and an efficient method is provided to compute the convolution results of the secondary filters by investigating their connection with the primary filters. Experiments on benchmarks demonstrate that, equipping CNNs with our versatile filters can lead to lower memory and computational cost, but with comparable network accuracy.

A preliminary version of this work was presented earlier~\cite{Versatile}. The present work adds to the initial version in significant ways. First, beyond the hand crafted binary masks, we propose to learn binary masks for different primary filters. The learned binary masks can be deployed in two kinds of ways: binary masks shared by different primary filters or different primary filters have distinct sets of binary masks. Second, to promote the diversity of generated secondary filters, we introduce orthogonal constraints to decrease the correlation between the learned binary masks. Third, we detail the optimization of primary filers and the learnable binary masks. To speed up the convolutions of versatile filters, we take convolutions between primary filters and input feature maps as intermediate results in the cache that can be repeatedly used further by different binary masks. In addition, extended experiments on different visual recognition tasks such as image classification, object detection and image super-resolution are conducted to verify the effectiveness of the versatile filters.

This paper is organized as follows. Section 2 revisits the related works on model compression and efficient operation design. The proposed versatile filters including spatial versatile filters, channel versatile filters and versatile filters with learnable masks, are described in Section 3 and Section 4. Section 5 shows the detailed experiments on various visual tasks to verify the effectiveness of the proposed method. Section 7 gives the conclusion of this paper.

\section{Related Work}
In this section, we will revisit these existing works from two aspects: model compression and efficient operation design.

\subsection{Model Compression}
\textbf{Weight Decomposition.}
The convolutional neural networks mainly consist of convolutional layers and fully connected layers. The number of weights in both  convolutional layers and fully connected layers is usually large in modern CNNs, and consequently the computational cost is considerably large. The weight decomposition methods aim to reduce the redundancy in the huge number of parameters by discovering the intrinsic correlations between weights~\cite{SVD,matrix1,fsnet,savarese2019learning}. Denton \etal~\cite{SVD} utilized matrix singular value decomposition (SVD) algorithm to generate a low-rank approximation of the weight matrix. Similarly in convolutional layers, Lebedev \etal~\cite{matrix3} computed a low-rank CP-decomposition of the weight tensor into the sum of a
small number of rank-1 tensors. Jaderbergn \etal~\cite{filterbank} approximated a full-rank filter bank as combinations of rank-1 filter bases in the spatial domain. Liu~\cite{sparseNet} performed sparse and low-rank decomposition of the weights and proposed an efficient sparse matrix multiplication for model acceleration. However, the rank of weights might be large in practice due to the need to extract diverse information of convolutional filters. Utilizing a rather low-rank approximation of the weights will significantly decrease the performance of the original models. FSNet \cite{fsnet} proposes a novel representation of filters, termed Filter Summary (FS), to enforce weight sharing across nearby filters. FS is a 1D vector from which filters are extracted as overlapping 1D segments. Beyond nearby filters in FSNet, we define and optimize binary marks to produce seconder filters in a more flexible way.

\textbf{Network Pruning.}
Pruning aims to cut out the less important connections or filters to reduce the redundancy and complexity of neural networks~\cite{pruning15,pruning,CNNpackNIPS,Thinet,tang2020scop}. Deep Compression \cite{pruning} integrates the element-wise pruning, nonlinear quantization and Huffman coding for network compression. Though its compression ratio is high, the inference speed-up cannot be easily achieved in practice, because of the hardware-unfriend element-wise pruning. During the inference phase, dequantization and Huffman decoding are necessary to recover the weights processed by Nonlinear quantization and Huffman coding, which further hurts the inference efficiency. Wang~\etal~\cite{CNNpackNIPS} explored to prune the connections in the frequency domain. Such connection pruning methods produce non-structured connectivity which adversely impacts practical acceleration in hardware platforms. Wen~\etal~\cite{sparse2} directly learned a compressed structure of CNNs by group Lasso regularization on the structures (\ie,~filters, channels, filter shapes, and layer depth) during the training. Network slimming in~\cite{liu2017learning} pruned the channels depending on channel scaling factors with the sparsity regularization during training. Pruning is simple and easy to obtain practical model compression and acceleration, but directly removing parameters and filters without other compensation limits its compression ratio.

\textbf{Weight Quantization.}
The values of weights and activations in neural networks are usually represented as 32-bit floating-point numbers. Model quantization represents the weights or the activations with low-bit integer numbers so as to reduce the memory usage and computational cost. Vanhoucke \etal~\cite{quan1} utilized fixed-point instructions to optimize the floating-point baseline. Wu \etal~\cite{wu2016quantized} proposed to train the neural networks directly with 8-bit integer quantized weights and activations. Zhang \etal~\cite{lqnets} proposed to jointly train a quantized, bit-operation-compatible DNN and its associated quantizers. Yang \etal~\cite{yang2019quantization} used a continuous function to approximate the quantization function by progressively making the continuous function sharper. Specifically, binarization methods~\cite{binaryconnect,bnn,xnor,yang2020searching,bireal,han2020training} with only 1-bit values can extremely accelerate the model by efficient binary operations. However, the low-bit quantization often leads to a significant accuracy drop compared to the original CNN.

\textbf{Knowledge Distillation.}
Knowledge distillation, also known as teacher-student framework, aims to transfer knowledge from \emph{teacher} models to \emph{student} models, where student models usually enjoy a lighter architecture than that of teacher models. Hinton \etal~\cite{Distill} first proposed the concept of knowledge distillation by introducing the teacher's softened output. Romero \etal~\cite{fitnets} further distilled the knowledge underlying the features of the teacher model to the student model. Chen~\etal~\cite{chen2019data} proposed to distill the student network without provided data. However, the major aim of knowledge distillation is to improve the performance of student models rather than designing compact networks.

In summary, pruning methods remove the unimportant weights or filters, while the proposed versatile filters generate subfilters from the intrinsic filters. Quantization methods represent the weights or activations in low-bit values, while our method operates at the filter level. Compared with the related works on extracting filters during inference, \eg, FSNet \cite{fsnet} enforces weight sharing across nearby filters and Savarese~\etal~\cite{savarese2019learning} shares weights across layers, the proposed versatile filters generate subfilters by hand-designed or learnable patterns.

\subsection{Efficient Operation Design}
Although model compression can significantly reduce the computational and memory cost, the performance of the compressed model is often upper bounded by that of the original neural networks. Designing more efficient operations for neural networks can bring the new potential for better models. Recently, a series of efficient operations are proposed for building lightweight neural networks~\cite{squeezenet,xception,mobilenet,mobilev2,mobilenetv3,shufflenet,shufflev2,igc,ghostnet}. To reduce the complexity of the convolution operation, Xception~\cite{xception} proposed to utilize depthwise convolution and pointwise convolution to approximate ordinary convolution. MobileNets~\cite{mobilenet} further stacked depthwise separable convolutions to form the lightweight deep neural networks.
MobileNetV2~\cite{mobilev2} upgraded the residual block~\cite{he2016deep} to the inverted residual block with pointwise convolutions and depthwise convolutions. ShuffleNet~\cite{shufflenet} was another lightweight network which introduces channel shuffling to handle the information exchange in group convolution. ShuffleNetV2~\cite{shufflev2} considered the memory usage and inference time in actual hardware for improving the efficiency of ShuffleNet. OctConv~\cite{chen2019drop} mixes different spatial sizes for different channels to make receptive field versatile. MixConv~\cite{mixnet} mixes different kernel sizes to obtain different receptive fields for different channels.
These efficient operations have made great progress for building more compact models. In this paper, we treat convolutional filters from an additive perspective by generating more secondary filters from a few primary filters.

\section{Spatial and Channel Versatile Filters}\label{Sec:indcompress}
In this section, we illustrate the design of versatile filters, which can be applied over any filter with height and width greater than one. Besides spatial versatile filters, we additionally investigate versatile filters from the channel perspective.

\subsection{Spatial Versatile Filters}\label{Sec:MSConv}
Consider the input data $x\in\mathbb{R}^{H\times W\times c}$, where $H$ and $W$ are height and width of the input data respectively, and $c$ is the channel number, \ie,~the number of feature maps generated in the previous layer. A convolution filter is denoted as $f\in\mathbb{R}^{d\times d\times c}$, where $d\times d$ is the size of the convolution filter. We focus on square filters, \eg,~$5\times5$ and $3\times3$, which are most widely used in modern CNNs such as ResNet~\cite{he2016deep}, VGGNet~\cite{VGGnet}, ResNeXt~\cite{ResNeXt}, and ShuffleNet~\cite{shufflenet}. The conventional convolution can be formulated as
\begin{equation}
	y = f\ast x,
	\label{Fcn:Conv}
\end{equation}
where $\ast$ is the convolution operation, $y\in\mathbb{R}^{H'\times W'}$ is the output feature map of $x$, and $H'$ and $W'$ are its height and width, respectively. 

Compared with traditional fully connected neural networks, one of the most important advantages of CNNs is that the size ($d\times d$) of filters in a convolutional layer can be much smaller than that ($H\times W$) of the input. For example, $7\times 7$ filters in the first layer of ResNet-50~\cite{he2016deep} are used to process  the $224\times 224$ input. Fixing the output size, the complexity of floating number multiplications of a filter in the fully-connected layer is $\mathcal{O}(cHWH'W')$, while the complexity of a convolution filter is only $\mathcal{O}(d^2cH'W')$. In addition,  convolution operations extract features from small regions, which is beneficial for subsequent tasks such as recognition and detection.

Receptive field is an important concept introduced by convolutions. The larger receptive field would allow neurons to detect changes over a wider area, but result in a less precise perception. On the other hand, the smaller receptive field would enable neurons to detect fine details. It is therefore reasonable to integrate neurons of larger receptive fields and smaller receptive fields to extract comprehensive and accurate features. For example, versatile modules \cite{GoogleNet} introduce parallel paths with different receptive field sizes by making use of multiple filters with different sizes, \eg, $3\times 3$ and $5\times 5$ convolutions. Explicitly bringing in filters of different sizes is a straightforward approach to process the input information in different scales, but the significant increase in storage of these filters could be a new challenge. Most importantly, though filters of different sizes in the same layer have different receptive fields, their receptive fields would have some overlap, which indicates the prospective connections between their corresponding filters.   

\begin{figure}[t]
	\centering
	\includegraphics[width=1.0\linewidth]{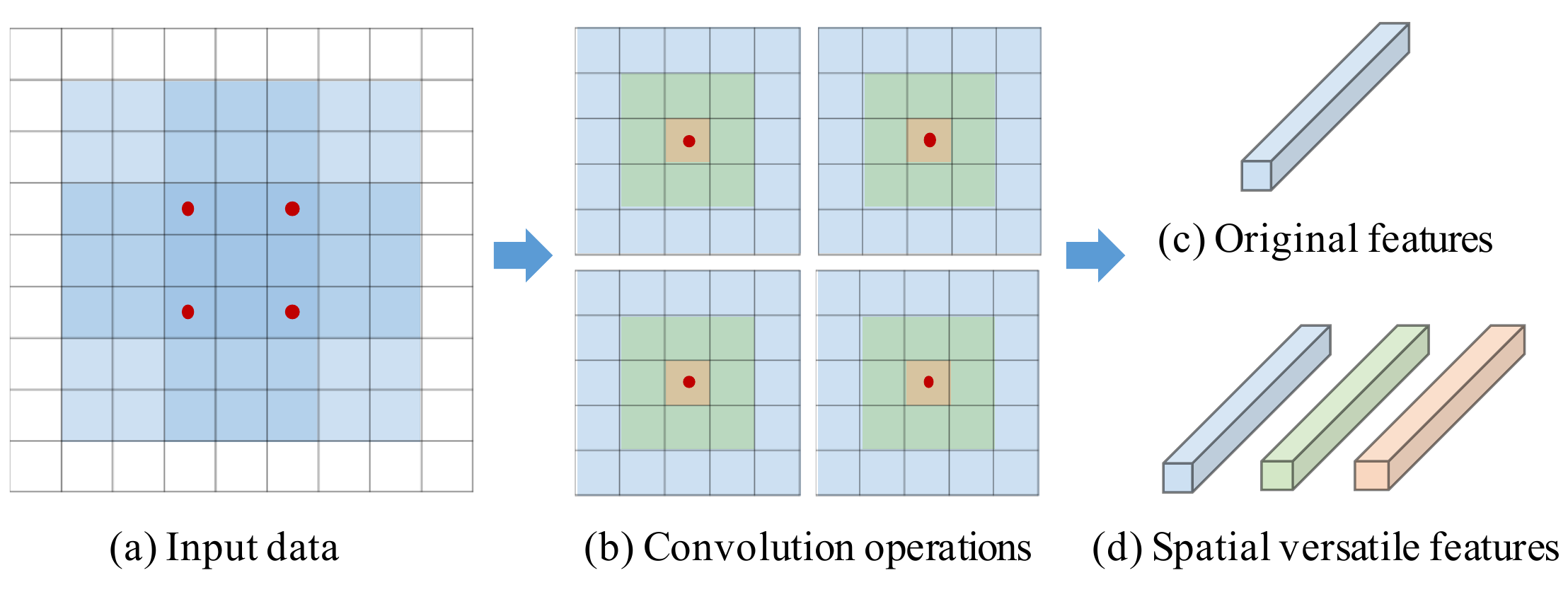}
	\caption{An illustration of the proposed spatial versatile convolution filter. Given the input data (a), there are four sub-regions (b) covered by a $5\times5$ convolution filter with stride 2, and their convolution results are stacked into a feature map (c). In contrast, a spatial versatile filter will be applied three times on each sub-region with different secondary filters, \ie,  $5\times5$ blue, $3\times3$ green, and $1\times1$ red in (b) to generate three feature maps (d).}
	\label{Fig:diagram}
\end{figure}

Taking $f\in\mathbb{R}^{d\times d}$ as a primary filter, we propose to derive a series of secondary filters $\{\hat f_{1}, \hat f_{2}, \cdots, \hat f_{s}\}$ from $f$, where $s = \lceil d/2 \rceil$. To maximally explore the potential of primary filter $f$, each secondary filters $\hat f_{i}$ is directly inherited from $f$ with a mask $M_{i}$,
\begin{equation}
	M_i(p,q,c) = \bigg\{
	\begin{array}{lc}
		1, &if \;  q, p \geq i \; | \; p, q \leq d+1-i,\\
		0, & otherwise,\\
	\end{array}
	\label{Fcn:M}
\end{equation}
and $\hat f_{i}$ is calculated as $\hat f_{i}=M_i\circ f$, where $\circ$ is the element-wise multiplication. More specifically, $\hat f_1$ is the filter $f$ itself, $f_2$ discards the outermost circle of parameters in $f$, and $\hat f_s$ is the innermost circle of parameters in $f$ (\ie, $\hat f_s$ is a $1\times1$ filter given an odd $d$). Example secondary filters for a $5\times 5$ filter can be seen in Figure~\ref{Fig:diagram} (b). 

By concatenating convolution responses from these secondary filters, we get the feature map represented as
\begin{equation}
	\centering
	\begin{aligned}
		y = &\left[(M_1\circ f)\ast x+b_1,\cdots,(M_s\circ f)\ast x+b_s\right],\\
		&s.t.\;\; s = \lceil d/2 \rceil,\;\; \{M_{i}\}_{i=1}^s\in\{0,1\}^{d\times d\times c},
	\end{aligned}
	\label{Fcn:MSConv}
\end{equation}
where $b_1,\cdots,b_s$ are bias parameters.

By embedding Eq.~\ref{Fcn:MSConv} into conventional CNNs, we can obtain convolution responses simultaneously from $s$ secondary filters of different receptive fields. The number of the output channels of the proposed versatile filter is $s$ times more than that of the original filter, and feature maps of a convolutional layer using the proposed versatile filters contain features in different scales at the same time.

Note that convolution operations ($\ast$) in Eq.~\ref{Fcn:MSConv} share the same stride and padding parameters for the following two reasons: 1) dimensionalities of feature maps generated by secondary filters with different receptive fields have to be consistent for the subsequent calculation; 2) centers of these secondary filters are the same, and the $s$-dimensional feature is thus a multi-scale representation of a specific pixel at $x$. Compared with original convolution filters, the proposed spatial versatile filters can provide more feature maps without increasing the number of filters. Therefore, the memory usage and computation cost of neural networks using the proposed spatial versatile filters can be significantly reduced. 

\noindent\textbf{Discussion:} Besides the proposed method as shown in Eq.~\ref{Fcn:MSConv}, a na\"{\i}ve approach to aggregate features from multiple secondary filters can be
\begin{equation}
	\centering
	\begin{aligned}
		y = &\sum_{i=1}^s(M_i\circ f)\ast x+b,\\
		s.t.\;\;  s = \lceil &d/2 \rceil,\;\; \{M_{i}\}_{i=1}^s\in\{0,1\}^{d\times d\times c},
	\end{aligned}
	\label{Fcn:MSconv2}
\end{equation}
which calculates the resulting feature map as a linear combination of features from different receptive fields. Since the convolution $\ast$  is exactly an linear operation, the sum of different convolution responses on the same input can be rewritten as the response of a combined convolution filter employed on this data, \ie,
\begin{equation}
	y = \sum_{i=1}^s(M_i\circ f)\ast x+b=[(\sum_{i=1}^sM_i)\circ f]\ast x+b.
\end{equation}
Therefore, Eq.~\ref{Fcn:MSconv2} is equivalent to adding a fixed weight mask on conventional convolution filters, which cannot produce more meaningful calculations and informative features in practice. We will compare the performance of this na\"{\i}ve approach in experiments.

To optimize the parameters in versatile filters, the standard back-propagation scheme can be applied accordingly for calculating the gradients. In addition, both convolution filters ${\mathbf f}$ and input data $\mathbf X$ are utilized multiple times in the proposed method, and their gradients are aggregated from different scales, which is $s\times$ larger than those in the conventional CNNs. Therefore, we divide them by the number of scales $s$ to avoid the gradient explosion in practice, and filters will be updated according to the learning rate $\eta$, \ie,~$\mathbf f = \mathbf f-{\partial\mathcal{L}}/{\partial\mathbf f}$.

\begin{figure}[t]
	\centering
	\includegraphics[width=1.0\linewidth]{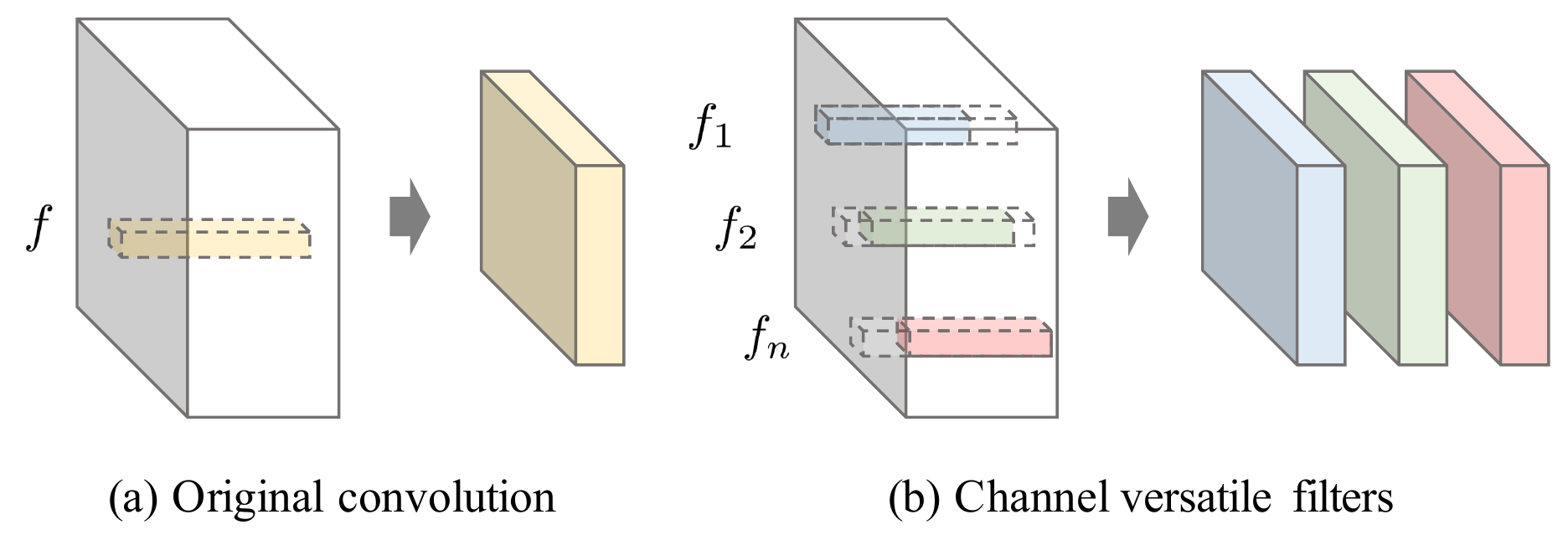}
	\caption{ An illustration of the proposed channel versatile filters. The original filter can generate only one feature map for the given input data, and the proposed method can provide multiple feature maps simultaneously according to the channel stride parameters. Each color represents a secondary filter and its corresponding feature map.}
	\label{Fig:diagram2}
\end{figure}

\subsection{Channel Versatile Filters}\label{Sec:Channel}
A spatial versatile filter was proposed in Eq.~\ref{Fcn:MSConv}, which generates a series of secondary convolution filters by adjusting the height and width of a given convolution filter. However, there is still obvious redundancy in these secondary filters, \ie, the number of channels of each convolution filter is much larger than its height and width. In addition, given $1\times 1$ primary filters, Eq.~\ref{Fcn:MSConv} will be reduced to Eq.~\ref{Fcn:Conv} for conventional convolution operation. Considering the wide use of $1\times1$ filters in modern CNN architectures such as MobileNet~\cite{mobilenet} and ShuffleNet~\cite{shufflenet}, we proceed to develop versatile filters from the channel perspective.

The most important property of convolution filters is that their weights are shared by the input data. A convolution filter used to have the same depth as the input data, and slide along the width and height of the input data with some stride parameters. If the depth of the input is 512, a $1\times1\times512$ filter has to take a large number of floating-point multiplications to weight different channels and integrate the information across different input channels. However, this coarse information summarization over all channels is difficult to highlight characters of individual channels, especially when there are extremely many channels. Hence, we define secondary filters $f_{i}$ for original convolution filters with the help of channel stride, \ie,
\begin{equation}
	\centering
	\begin{aligned}
		y = &\left[\hat f_{1}\ast x_{1},\hat f_{2}\ast x_{2},\cdots,\hat f_{n}\ast x_{n}\right],\\
		&s.t.\;\; \forall~i, \; \hat f_{i}\in\mathbb{R}^{d\times d\times c}, \;\; n = (c-\hat{c})/g+1.
	\end{aligned}
	\label{Fcn:Channel}
\end{equation}
where $g$ is the channel stride parameter, $\hat{c}<c$ is the number of non-zero channels of secondary filters, and the bias term is omitted for simplicity. $\hat f_{i}$ is a subset of primary filter $f$ in the $i$-th sliding window with the length $\hat{c}$ and the stride $g$. Therefore, a filter will be used $n$ times simultaneously to generate more feature maps by introducing Eq.~\ref{Fcn:Channel}. Example secondary filters using the proposed channel stride approach are given in Figure~\ref{Fig:diagram2}. In addition, the proposed channel versatile filters can also significantly reduce the memory usage and computational complexity of CNNs, which is further analyzed in Section \ref{sec:complexity}.

\section{Learning to Be Versatile}

\begin{figure*}[t]
	\centering
	\begin{tabular}{cc}
		\includegraphics[width=0.4\linewidth]{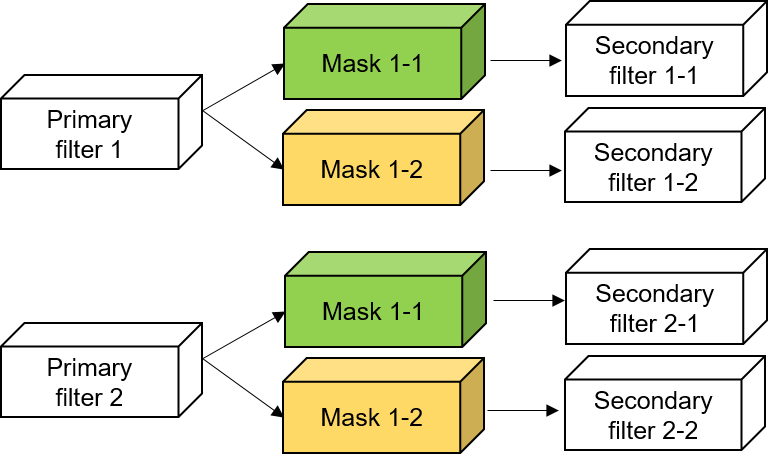} \hspace{0.2cm} &\includegraphics[width=0.4\linewidth]{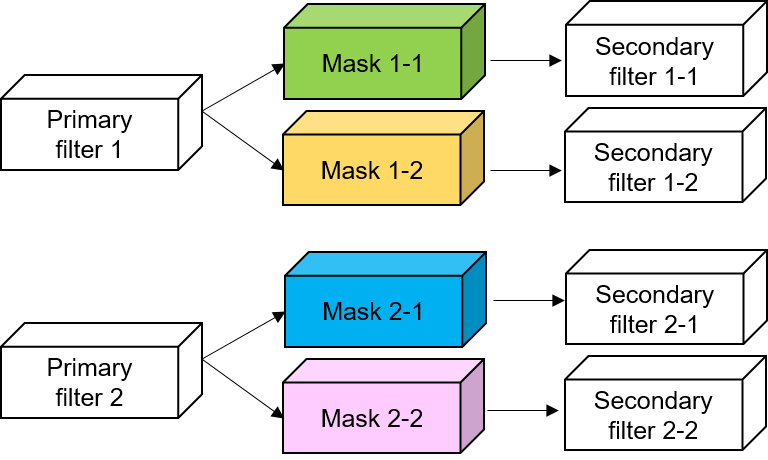}
		\\
		\small (a) The secondary filters generation using shared masks. & \small (b) The secondary filters generation using separate masks.
	\end{tabular}
	\caption{An illustration of the proposed versatile filters method for generating secondary filters using primary filters and different mask strategies, \ie, shared and separate associated masks, which can be selected according to the practical requirement. Wherein, associated masks in the same color are exactly the same one. Better view in the color version.}
	\label{Fig:illustration}
	\vspace{-1em}
\end{figure*}

Given a primary filter $f\in\mathbb{R}^{d\times d\times c}$ and a binary mask $M_i\in\{0,1\}^{d\times d\times c}$, a secondary filter can be generated through $\hat f_i=f\circ M_i$, as shown in Eq. (\ref{Fcn:MSConv}). However, these hand-crafted masks are highly related to human knowledge and may not always be optimal for CNNs in various visual recognition tasks.

Instead of pre-defining the rule to produce the binary mask, we tend to learn the optimal binary masks along with the optimization of the deep neural network. If there are $s$ different binary masks, then we can generate $s$ secondary filters $\{\hat f_i\}_{i=1}^s$ from a primary filter $f$. Suppose that we learn $k$ primary filters and  $s$ binary masks to generate a similar number of convolution filters ($k\times s \approx n$) as the original neural network. A primary filter will be exploited for many times (up to $s$) in the proposed method, but only one copy has to be maintained, which implies that fewer 32-bit values are required than that of standard convolution filters. Since binary masks are not fixed and will vary with different primary filters, we need to store them as well, but 1-bit binary values would not bring in too much storage cost. 

When there are multiple primary filters, we propose two strategies for deploying binary masks to different primary filters, namely, the \textit{shared} strategy and the \textit{separate} strategy. In the shared strategy, the binary masks are shared with different primary filters, while the masks are separate for each primary filter in the separate strategy. In particular, the different primary filters and the same mask can generate different secondary filters in the shared strategy, \eg,~$\hat{f}_{11}$ and $\hat{f}_{21}$ are secondary filters generated by applying the same mask $M_1$ on the first and the second primary filters, as shown in Figure~\ref{Fig:illustration}(a). Formally, 
\begin{equation}\label{eq:share}
	\hat{f}_{ij}=f_i\circ M_{j},
\end{equation}
and the convolution in the shared strategy can be written as
\begin{equation}\label{eq:conv-share}
	\begin{aligned}
		y &= [\hat{{f}}_{11}*x,\hat{{f}}_{12}*x,\cdots,\hat{{f}}_{ks}*x]\\
		&= [(f_1\circ M_{1})*x,(f_1\circ M_{2})*x,\cdots,(f_k\circ M_{s})*x].
	\end{aligned}
\end{equation}
The masks are binary and would not account for an obvious proportion of the entire memory usage of CNNs by the proposed method. It is therefore applicable to assign each primary convolution filter with a separate set of $s$ binary masks, so that we can harvest abundant diverse secondary filters at a small memory cost. To this end, we further extend Eq. \ref{eq:share} to contain $k\times s$ binary masks, and each secondary filter is represented as
\begin{equation}\label{eq:ind}
	\hat{f}_{ij}=f_i\circ M_{ij},
\end{equation}
where $i=1,\cdots,k$, and $j=1,\cdots,s$. The convolution operation using separate mask sets can be formulated as
\begin{equation}\label{eq:conv-ind}
	\begin{aligned}
		y &= \![\hat{{f}}_{11}*x,\hat{{f}}_{12}*x,\cdots,\hat{{f}}_{ks}*x]\\
		&= \![(f_1\!\circ\! M_{11})*x,(f_1\!\circ\! M_{12})*x,\cdots,(f_k\!\circ\! M_{ks})*x],
	\end{aligned}
\end{equation}
whose computational complexity is the same as that of the shared strategy. Although the number of masks is increased, this separate strategy can lead to better representation capability, because of the increased diversity of generated secondary filters. 

To further promote the diversity of the generated secondary filters and their resulting feature maps, binary masks for a primary filter should be different from each other as much as possible. Therefore, we propose to decrease the correlation between binary masks. By vectorizing and stacking all $s$ binary masks into a matrix, we have
\begin{equation}
	\mathbf{M} = [\vect(M_1),\cdots,\vect(M_s)],
\end{equation}
where $\mathbf{M}$ should be an approximately orthogonal matrix for generating feature maps without obvious correlation. We therefore apply the following regularization on binary masks in each convolutional layer during the training stage:
\begin{equation}
	\mathcal{L}_{\text{ortho}} = \frac{1}{2}\|\frac{1}{d^2c}\mathbf{M}^T\mathbf{M}-\mathbf{I}\|_F^2,
\end{equation}
where $\mathbf{I}\in\mathbb{R}^{s\times s}$ is the identity matrix, $\|\cdot\|_F$ is the Frobenius norm for matrices, and $d^2c$ is the number of elements in each column of $\mathbf{M}$. By minimizing the above function, binary masks would be approximately orthogonal with each other, which will improve the difference between secondary filters and increase the informativeness of resulting feature maps.

\subsection{Optimization}
The convolution operation in CNNs can be reformulated as the matrix multiplication, and the input data $x$ of a given sample is usually divided into several overlapping areas in practice. Dividing the input data $x$ into $l = H'\times W'$ areas (the size of each area is $d\times d\times c$) and vectorizing them, we have $\mathbf X = [\text{vec}(x_1),\cdots,\text{vec}(x_l)]\in\mathbb{R}^{d^2c\times l}$. Denote $\hat{\mathbf{F}}=[\vect(\hat f_1),\cdots,\vect(\hat f_n)]=[\hat{\mathbf{f}}_1,\cdots,\hat{\mathbf{f}}_n]\in\mathbb{R}^{(d^2c)\times n}$, we can rewrite the convolution (Eq.~\ref{Fcn:Conv}) in matrix form:
\begin{equation}\label{eq:conv2}
	\mathbf{Y} = [\mathbf{X}^T\hat{\mathbf{f}}_1,\cdots,\mathbf{X}^T\hat{\mathbf{f}}_n],
\end{equation}
where the $i$-th column in $\mathbf{Y}\in\mathbb{R}^{l\times n}$ is the convolution result of all the patches using the $i$-th filter in $\hat{\mathbf{F}}$. Here we introduce how to learn the primary filters and binary masks. Shared and separate strategies are designed to deploy the binary masks for versatile filters. Here we take the shared strategy as an example to illustrate the optimization. There are two sets of learnable variables, \ie,~the primary filters  $\mathbf{F}=[\mathbf{f}_1,\cdots,\mathbf{f}_k]\in\mathbb{R}^{(d^2c)\times k}$ and the  binary masks $\mathbf{M}=[\mathbf{m}_1,\cdots,\mathbf{m}_s]\in{\{0,1\}}^{(d^2c)\times s}$. The objective function of the proposed algorithm can be written as
\begin{equation}\label{eq:object}
	\min_{\mathbf{F},\mathbf{M}} \mathcal{L} = \mathcal{L}_{0}(\mathbf{F},\mathbf{M}) + \lambda\mathcal{L}_{\text{ortho}}(\mathbf{M}),
\end{equation}
where $\mathcal{L}_{0}$ indicates a loss function for the task at hand, \eg, cross entropy loss for classification task and mean square error loss for regression task, and $\lambda$ is the hyper-parameter for orthogonal regularization.

\textbf{Updating $\mathbf{F}$.}
The gradient of the second term in the object function can be easily obtained. In particular, the orthogonal regularization is rewritten as
\begin{equation}\small
	\begin{aligned}
		\mathcal{L}_{\text{ortho}} &= \frac{1}{2}\text{Tr}\left[ \left(\frac{1}{d^2c}\mathbf{M}^T\mathbf{M}-\mathbf{I}\right)\left(\frac{1}{d^2c}\mathbf{M}^T\mathbf{M}-\mathbf{I}\right)^T \right] \\
		&= \frac{1}{2}\text{Tr}\left( \frac{1}{d^4c^2}\mathbf{M}^T\mathbf{M}\mathbf{M}^T\mathbf{M} - 2\frac{1}{d^2c}\mathbf{M}^T\mathbf{M} + \mathbf{I} \right),
	\end{aligned}
\end{equation}
whose gradient \wrt $\mathbf{M}$ is given by
\begin{equation}
	\begin{aligned}
		\frac{\partial \mathcal{L}_{\text{ortho}}}{\partial \mathbf{M}} &= \frac{1}{2}\left( \frac{4}{d^4c^2}\mathbf{M}\mathbf{M}^T\mathbf{M} - \frac{4}{d^2c}\mathbf{M} \right) \\
		&= \frac{2}{d^4c^2}\mathbf{M}\mathbf{M}^T\mathbf{M} - \frac{2}{d^2c}\mathbf{M}.
	\end{aligned}
\end{equation}
For the first term in Eq.~\ref{eq:object}, both gradients \wrt $\mathbf{F}$ and $\mathbf{M}$ should be calculated. During the gradient back-propagation, the gradient $\frac{\partial \mathcal{L}_0}{\partial \hat{\mathbf{f}}_{ij}}$ of loss function $\mathcal{L}_0$ \wrt the generated secondary filter $\hat{\mathbf{f}}_{ij}$ can be obtained according to the standard back-propagation rule. The gradient to $\mathbf{f}_i$ is calculated by
\begin{equation}
	\frac{\partial \mathcal{L}_0}{\partial \mathbf{f}_i} = \sum_{j=1}^{s}\frac{\partial \mathcal{L}_0}{\partial \hat{\mathbf{f}}_{ij}}\circ \mathbf{m}_j,
\end{equation}
and the gradient to individual mask $\mathbf{m}_j$ is
\begin{equation}
	\frac{\partial \mathcal{L}_0}{\partial \mathbf{m}_j}=\sum_{i=1}^{k}\frac{\partial \mathcal{L}_0}{\partial \hat{\mathbf{f}}_{ij}}\circ \mathbf{f}_i.
\end{equation}
Then the overall gradients of the object function \wrt $\mathbf{F}$ and $\mathbf{M}$ are
\begin{equation}\label{eq:grad-b}
	\frac{\partial \mathcal{L}}{\partial \mathbf{F}} = [\frac{\partial \mathcal{L}_0}{\partial \mathbf{f}_1},\cdots,\frac{\partial \mathcal{L}_0}{\partial \mathbf{f}_k}],
\end{equation}
and 
\begin{equation}\label{eq:grad-m}
	\begin{aligned}
		\frac{\partial \mathcal{L}}{\partial \mathbf{M}} &= [\frac{\partial \mathcal{L}_0}{\partial \mathbf{m}_1},..,\frac{\partial \mathcal{L}_0}{\partial \mathbf{m}_s}] + \lambda\frac{\partial \mathcal{L}_{\text{ortho}}}{\partial \mathbf{M}},
	\end{aligned}
\end{equation}
respectively. By using stochastic gradient descent (SGD), $\mathbf{F}$ can be updated as
\begin{equation}\label{eq:b-update}
	\mathbf{F} \leftarrow \mathbf{F} - \eta\frac{\partial \mathcal{L}}{\partial \mathbf{F}},
\end{equation}
where $\eta$ is the learning rate. 

\textbf{Updating $\mathbf M$.}
Since elements in $\mathbf{M}$ are binary, the update $\mathbf{M}\leftarrow \mathbf{M}-\eta\frac{\partial \mathcal{L}}{\partial \mathbf{M}}$ cannot be directly applied. To make the optimization of the binary mask compatible with back-propagation, an agent variable $\mathbf{H}\in\mathbb{R}^{(d^2c)\times s}$ is introduced to represent $\mathbf{M}$ as 
\begin{equation}\label{eq:sgn}
	\mathbf{M} = \text{sign}(\mathbf{H}),
\end{equation}
where $\text{sign}(\cdot)$ is the sign function that outputs $+1$ for positive numbers and $0$ otherwise. 
In Eq. \ref{eq:sgn}, the derivative of the sign function is zero almost everywhere, which prevents the back-propagation process. Following ``straight-through estimator'' in~\cite{bengio2013estimating}, in forward propagation we simply apply sign function (Eq. \ref{eq:sgn}), but during backward propagation the gradient of $\mathbf{M}$ is back transmitted to $\mathbf{H}$ intactly:
\begin{equation}
	\frac{\partial \mathcal{L}}{\partial \mathbf{H}} = \frac{\partial \mathcal{L}}{\partial \mathbf{M}}.
\end{equation}
Then $\mathbf{H}$ is updated via 
\begin{equation}\label{eq:h-update}
	\left\{
	\begin{array}{ll}
		\mathbf{H} &\leftarrow~ \mathbf{M} \\
		\mathbf{H} &\leftarrow~ \mathbf{H} - \eta\frac{\partial \mathcal{L}}{\partial \mathbf{H}}.
	\end{array}
	\right.
\end{equation}
Since $\mathbf{H}$ is the hidden state of the binary mask, clipping is used to bound weights:
\begin{equation}
	\mathbf{H}\leftarrow \text{clip}\left(\mathbf{H}-\eta\frac{\partial \mathcal{L}}{\partial \mathbf{H}},0,1\right),
\end{equation}
where clipping is a common practice to bound weights in order to regularize them since sign function is not influenced by the magnitude of the real-value input and the $\mathbf{H}$ would otherwise grow very large with no change to the binary values. The update procedure for versatile filters with shared masks can be found in Alg.\ref{alg1}. The only difference of algorithm for separate strategy is that $\frac{\partial \mathcal{L}_0}{\partial \mathbf{m}_{ij}}=\frac{\partial \mathcal{L}_0}{\partial \hat{\mathbf{f}}_{ij}}\circ \mathbf{F}_i$.

\begin{algorithm}[t]
	\caption{Feed-Forward and Back-Propagation Process of Versatile Filters with Learnable Masks.} 
	\label{alg1}
	\begin{algorithmic}[1]
		\REQUIRE Random initialized primary filters $\mathbf{F}$, learnable masks $\mathbf{M}$, the agent variable $\mathbf{H}$, input feature map $X$, the loss function $\mathcal{L}$, the learning rate $\eta$. Note that $\mathbf{H}$ is only used during training.
		\STATE \textbf{Feed Foward:}
		\STATE If in training stage, obtain the binary masks: $\mathbf{M} = \text{Sign}(\mathbf{H})$ (Eq.\ref{eq:sgn});
		\STATE Compute the output feature maps: $\mathbf{Y} = [\mathbf{X}^T\hat{\mathbf{f}}_1,\cdots,\mathbf{X}^T\hat{\mathbf{f}}_n]$ (Eq.~\ref{eq:conv2});
		\STATE \textbf{Backward Propagation:}
		\FOR{$i=1,\cdots,k$ and $j=1,\cdots,s$}
		\STATE Calculate the gradient $\frac{\partial \mathcal{L}}{\partial \hat{\mathbf{f}}_{ij}}$ according to the standard back-propagation rule;
		\ENDFOR
		\STATE Compute the gradient of $\mathbf{F}$: $\frac{\partial L}{\partial \mathbf{F}}$ (Eq.\ref{eq:grad-b});
		\STATE Compute the gradient of $\mathbf{M}$: $\frac{\partial \mathcal{L}}{\partial \mathbf{M}}$ (Eq.\ref{eq:grad-m});
		\STATE Estimate $\frac{\partial \mathcal{L}}{\partial \mathbf{H}} = \frac{\partial \mathcal{L}}{\partial \mathbf{M}}$;
		\STATE \textbf{Parameter Update:}
		\STATE Update the masks' agent variable
		\vspace{-0.3cm}
		\begin{equation*}
			\left\{
			\begin{array}{ll}
				\mathbf{H} &\leftarrow~ \mathbf{M}, \\
				\mathbf{H} &\leftarrow~ \text{clip}\left(\mathbf{H}-\eta\frac{\partial \mathcal{L}}{\partial \mathbf{H}},0,1\right);
			\end{array}
			\right.
		\end{equation*}
		\vspace{-0.4cm}
		\STATE Update the primary filters $\mathbf{F} \leftarrow \mathbf{F} - \eta\frac{\partial \mathcal{L}}{\partial \mathbf{F}}$;
		\ENSURE Feature maps $\mathbf Y$, primary filters $\mathbf{F}$ and binary masks $\mathbf{M}$.
	\end{algorithmic}
\end{algorithm}

\subsection{Analysis on Complexities}\label{sec:complexity}
Compared with original convolution filters, the proposed versatile filters with handcrafted or learnable masks can provide more feature maps without increasing the number of filters. Therefore, we proceed to analyze the memory usage of versatile filters with binary masks, and provide an efficient implementation for the proposed convolutions, which theoretically brings in fewer multiplication operations.

\subsubsection{Model Size Compression}
The proposed versatile convolution operation can generate multiple feature maps using a fixed number of primary filters. Thus the computational complexity and memory usage of CNNs for extracting the same amount of features can be reduced significantly. Note that memory usage here refers to the size of storage. Given $n$ filters $\{f_i\}_{i=1}^{n}$ in a classical convolution layer, the number of parameters is $\mathcal{O}(d^2cn)$ 32-bit floating-point values. In spatial or channel versatile filters, we only need to store the parameters of primary filters, \ie, 
\begin{equation}
	\mathcal{O}(d^2cn/s).
\end{equation}
In versatile filters with shared learnable masks, to establish $n$ secondary filters of the size $d\times d\times c$, parameters of primary filters in the proposed convolution with shared binary masks are $\mathcal{O}(d^2cn/s)$ 32-bit floating-point values, and the shared masks have $\mathcal{O}(d^2cs)$ 1-bit binary values. The total space complexity for storing the parameters is
\begin{equation}
	\mathcal{O}(d^2cn/s + d^2cs/32).
\end{equation}
Given the separate strategy, the number of parameters in primary filters is the same, that is, $\mathcal{O}(d^2cn/s)$ 32-bit floating values. The separate masks need $\mathcal{O}(d^2cn)$ 1-bit binary values. Overall, the space complexity for the separate strategy is
\begin{equation}
	\mathcal{O}(d^2cn/s + d^2cn/32).
\end{equation}
It can be seen that versatile filters with shared masks lead to a more compact CNN of fewer parameters, while versatile filters with separate masks can increase the diversity of secondary filters and bring in richer representations of the input. The effects of versatile filters with both shared and separate strategies are thoroughly evaluated in the following experiments.

\subsubsection{Speeding Up Convolutions}\label{sec:mul}
In the conventional implementation of convolution operations \cite{jia2014caffe,tensorflow}, the input feature map is split into many patches and convolution is applied on each patch to compute the output feature map. Given a patch $x\in\mathbb{R}^{d\times d\times c}$ in the input feature map, the conventional convolution operation can be represented as
\begin{equation}
	\begin{aligned}
		y &= \left\{ f_i * x \right\}_{i=1}^{n} \\
		& = \left\{ \sum\left(\vect(x)\circ\mathbf{f}_i\right) \right\}_{i=1}^{n},
	\end{aligned}
\end{equation}
where $y\in\mathbb{R}^{1\times 1\times n}$. It involves $d^2cn$ multiplications (MUL) and $d^2cn$ add operations (ADD). For all of spatial versatile filters, channel versatile filters and versatile filters with learnable masks, we can conduct the convolution as
\begin{equation}\label{eq:old-mask}
	\begin{aligned}
		y &= \left\{ \sum\left( \vect(x)\circ\hat{\mathbf{f}}_{ij}\right) \right\}_{i=1,j=1}^{k,s}\\
		&= \left\{ \sum\left(\vect(x)\circ\mathbf{f}_i\circ\mathbf{m}_j \right)  \right\}_{i=1,j=1}^{k,s}.
	\end{aligned}
\end{equation}
It is instructive to note that $\vect(x)\circ \mathbf{f}_i$ has been repeatedly calculated for different masks $\mathbf{m}_j$. We thus only compute it for once and store it in cache. Given the intermediate result $\{\mathbf{c}_i=\vect(x)\circ \mathbf{f}_i\}_{i=1}^{k}$, Eq.~\ref{eq:old-mask} can be simplified as
\begin{equation}\label{eq:new-mask}
	y = \left\{ \sum\left(\mathbf{c}_i\circ \mathbf{m}_j\right)  \right\}_{i=1,j=1}^{k,s}.
\end{equation}
Noticing that $\mathbf{m}_j$ is binary, multiplication with it can be implemented efficiently by binary masking operation in practice, which takes much less time than standard multiplication. In total, the convolution in our method for one patch only needs $d^2ck$ FP32 MULs, $d^2cn$ ADDs and $d^2cn$ binary masking (MASK) operations. The computational complexity of binary masking is about 1/32$\times$ that of FP32 multiplication~\cite{xnor,binaryconnect}, so the combined number of MULs is calculated as
\begin{equation}
	\#MUL=\frac{1}{32}\#MASK+\#MUL_{fp32}.
\end{equation}
For the entire convolution process, the number of MUL operations for desiring feature map $y\in\mathbb{R}^{H'\times W'\times n}$ is about
\begin{equation}
	\mathcal{O}(d^2cH'W'k)=\mathcal{O}\left(d^2cH'W'n\left(\frac{1}{s}+\frac{1}{32}\right)\right).
\end{equation}
In Eq.~\ref{eq:new-mask}, a half of values on average in mask $\mathbf{m}_j$ is zero, so a half of additions can be reduced and the number of ADDs is about
\begin{equation}
	\mathcal{O}(0.5d^2cH'W'n).
\end{equation}
In order to maximize the speed-up effect of the multiplication reduction in versatile filters, more engineering efforts could be taken to further optimize the efficiency of versatile filters on various devices, \eg,~FPGA and NPU.

\section{Experiments}\label{Sec:Exp}
In this section, we will conduct experiments to validate the effectiveness of the proposed versatile filters on several benchmark image datasets, including CIFAR-10~\cite{cifar}, ImageNet (ILSVRC 2012~\cite{ImageNet}), MS COCO~\cite{lin2014microsoft} and Set5~\cite{set5}. Experimental results will be analyzed to further understand the benefits of the proposed approach. For concision, we denote CNNs constructed using spatial versatile filters, channel versatile filters and those versatile filters with learnable masks as S-Versatile, C-Versatile and L-Versatile, respectively.

\subsection{Experiments on CIFAR-10}
The CIFAR-10 dataset consists of $60,000$ images drawn from ten categories, which is split into $50,000$ training and $10,000$ validation images. Each example in this dataset is a $32\times32$ color image.

\subsubsection{Spatial Versatile Filters} We first tested the performance of the proposed spatial versatile filter in Eq.~\ref{Fcn:MSConv} using ResNet-56 as base network architecture for classifying the CIFAR-10 dataset~\cite{cifar}. All the models are trained using SGD optimizer for 400 epochs with weight decay of 5e-4. The initial learning rate is set as 0.1 and decays at 200-th, 300-th and 375-th epoch. ResNet-56 has 0.85M parameters, and achieves 93.5\% accuracy on CIFAR-10. The results of spatial versatile filter and some other state-of-the-art model compression methods are shown in Table~\ref{Tab:CIFAR}. Wherein, the number of parameters and floating-point multiplications of each model are also provided.

The proposed method can generate multiple feature maps using a convolution filter whose size is larger than $2\times2$ (\ie, $s = \lceil d_i/2 \rceil>1$), which will increase the number of channels in the next layer and make the convolutional neural network enormous. Therefore, we reduce the number of convolution filters in each layer to make the number of feature maps in S-Versatile model similar to that in the original network, as shown in Table~\ref{Tab:CIFAR}.
The performance of S-Versatile model is similar to that of the baseline model, but has significantly lower memory usage and FLOPs, which demonstrates the effectiveness of the proposed versatile convolution filters.

\begin{table}[h]
	\begin{center}
		\caption{The performance of the proposed spatial versatile filters on CIFAR-10. \#Param means the number of parameters counting in 32 bit, \#MUL means the number of combined multiplications, \#ADD means the number of addition operations, and Acc denotes the accuracy.}
		\label{Tab:CIFAR}
		\renewcommand{\arraystretch}{1.1} 
		\setlength{\tabcolsep}{4pt}{
			\begin{tabular}{l||c|c|c|c}
				\hline
				Model & \#Param & \#ADD & \#MUL & Acc\\
				\hline
				\hline
				Baseline & $8.5\times10^5$ &  $1.3\times10^8$ &  $1.3\times10^8$ & 93.5\%\\
				\hline
				L1 pruning~\cite{l1-pruning} &  $7.3\times10^5$ &  $0.9\times10^8$ &  $0.9\times10^8$ & 93.1\%\\
				NISP~\cite{nisp} & $7.1\times10^5$ &  $0.9\times10^8$ &  $0.9\times10^8$ & 93.0\% \\
				GAL-0.6~\cite{gal} & $7.5\times10^5$ &  $0.8\times10^8$ &  $0.8\times10^8$ & 93.0\% \\
				Channel Pruning~\cite{he2017channel} & - &  $0.6\times10^8$ &  $0.6\times10^8$ & 90.8\% \\
				HRank~\cite{hrank} & $4.9\times10^5$ &  $0.6\times10^8$ &  $0.6\times10^8$ & 93.2\% \\
				GAL-0.8~\cite{gal} & $2.9\times10^5$ &  $0.5\times10^8$ &  $0.5\times10^8$ & 90.4\% \\
				HRank~\cite{hrank} & $2.7\times10^5$ &  $0.3\times10^8$ &  $0.3\times10^8$ & 90.7\% \\
				\hline
				S-Versatile  & $4.3\times10^5$ &  $0.6\times10^8$ &  $0.6\times10^8$ & 93.2\%\\
				S+C-Versatile  & $2.2\times10^5$ &  $0.6\times10^8$ &  $0.3\times10^8$ & 92.1\%\\
				Shared L-Versatile (s=2)  & $4.3\times10^5$ & $0.6\times10^8$ &  $0.7\times10^8$ & 92.8\%\\
				Separate L-Versatile (s=2)  & $4.4\times10^5$ & $0.6\times10^8$ &  $0.7\times10^8$ & 93.4\%\\
				Separate L-Versatile (s=4)  & $2.2\times10^5$ & $0.6\times10^8$ &  $0.4\times10^8$ & 92.8\%\\
				\hline
			\end{tabular}
		}
	\end{center}
\end{table}

\subsubsection{Channel Versatile Filters} We further test the performance of the proposed channel versatile filters as described in Eq.~\ref{Fcn:Channel}, namely C-Versatile. Note that, for the first layer and the last layer in neural networks, we do not apply the channel stride approach, since the input channel of the first layer is usually very small and the output channel of the last layer is exactly the number of ground-truth labels. There are two important parameters in Eq.~\ref{Fcn:Channel}, \ie, the number of channels $\hat{c}$ of the convolution filter $\hat{f}$ and the stride $g$. We then established three models using the proposed versatile filter with different $\hat{c}$ and $g$, and trained them on the CIFAR-10 dataset as detailed in Table.~\ref{Tab:CIFAR}. 

As mentioned above, the channel versatile filters can reduce the number of convolution filters by a factor of $n = (c-\hat{c})/g+1$, therefore, when we set $\hat{c}-c=g$, we can reduce about half convolution filters and maintain the similar amount of feature maps. $\hat{c}-c$ determines the overlap ratio and the input channels of sub-filters. If $\hat{c}-c$ is small, the overlap is much and diversity will be less, and otherwise the input channels will be small. From Table~\ref{Tab:CIFAR2}, C-Versatile with $\hat{c}-c=g=8$ achieved the best accuracy, \ie,~$93.4\%$ accuracy, which is slightly lower than that of the baseline model, but its memory usage and FLOPs have been reduced significantly. Furthermore, when $(\hat{c}-c)/g>1$, the number of convolution filters will be further reduced. However, since the number of filters is very small, the representability of this network is also lower. Therefore, we set $\hat{c}-c=8$ and $g=8$ in the following experiments for having a best trade-off. We also integrate the spatial and the channel versatile filters on ResNet-56, and show the result in Table~\ref{Tab:CIFAR}. The model is compressed by about 4$\times$ and still obtain $92.1\%$ accuracy.

\begin{table}[h]
	\begin{center}
		\caption{The performance of the proposed channel versatile filters on CIFAR-10.}
		\label{Tab:CIFAR2}
		\renewcommand{\arraystretch}{1.1}
		\setlength{\tabcolsep}{8pt}
		\begin{tabular}{l||c|c|c|c|c}
			\hline
			Model & $c$-$\hat{c}$ & g & \#Param &  \#MUL &Acc\\
			\hline
			\hline
			Baseline & - & - & $8.5\times10^5$ &  $1.3\times10^8$ & $93.5\%$\\
			\hline
			C-Versatile & 1 & 1 & $4.2\times10^5$& $0.6\times10^8$ & $92.4\%$ \\
			C-Versatile & 2 & 2 & $4.2\times10^5$ & $0.6\times10^8$ & $92.7\%$ \\
			C-Versatile & 4 & 4 & $4.2\times10^5$ & $0.6\times10^8$ & $93.0\%$ \\
			C-Versatile & 8 & 8 & $4.2\times10^5$ & $0.6\times10^8$ & $93.4\%$ \\
			C-Versatile & 12 & 12 & $4.2\times10^5$ & $0.6\times10^8$ & $92.4\%$ \\
			\hline
			C-Versatile & 8 & 4 & $2.8\times10^5$&  $0.4\times10^8$ &  $92.4\%$\\
			C-Versatile & 8 & 2 & $2.2\times10^5$&  $0.3\times10^8$ &  $91.2\%$\\
			\hline
		\end{tabular}
	\end{center}
\end{table}

\subsubsection{Versatile Filters with Learnable Masks}
We replace the conventional convolution filters using versatile filters with learnable masks for all the convolutional layers, except the first layer. When we set $s=2$, the results in Table \ref{Tab:CIFAR} show that both Shared L-Versatile and Separate L-Versatile can obtain comparable performance with the baseline, meanwhile yielding about $2\times$ and $1.9\times$ compression ratios, respectively. Further increasing $s$ to 4 lowers performance slightly but the memory has been reduced significantly and \#MUL is reduced by $s\times$. On the other side, Separate L-Versatile performs better than Shared L-Versatile consistently. Compared to the shared strategy, the separate strategy can provide more diverse masks and improve the capacity of the network.

To justify the advantages of the learned binary masks, we also test versatile filters with random fixed binary masks (Shared L-Versatile-Rand and Separate L-Versatile-Rand in Table \ref{Tab:CIFAR}). The results in Table \ref{Tab:CIFAR} show that the learnable masks enjoy more performance improvement.

\begin{figure}[t]
	\setlength{\tabcolsep}{1mm}
	\begin{tabular}{cc}
		\includegraphics[width=0.24\textwidth]{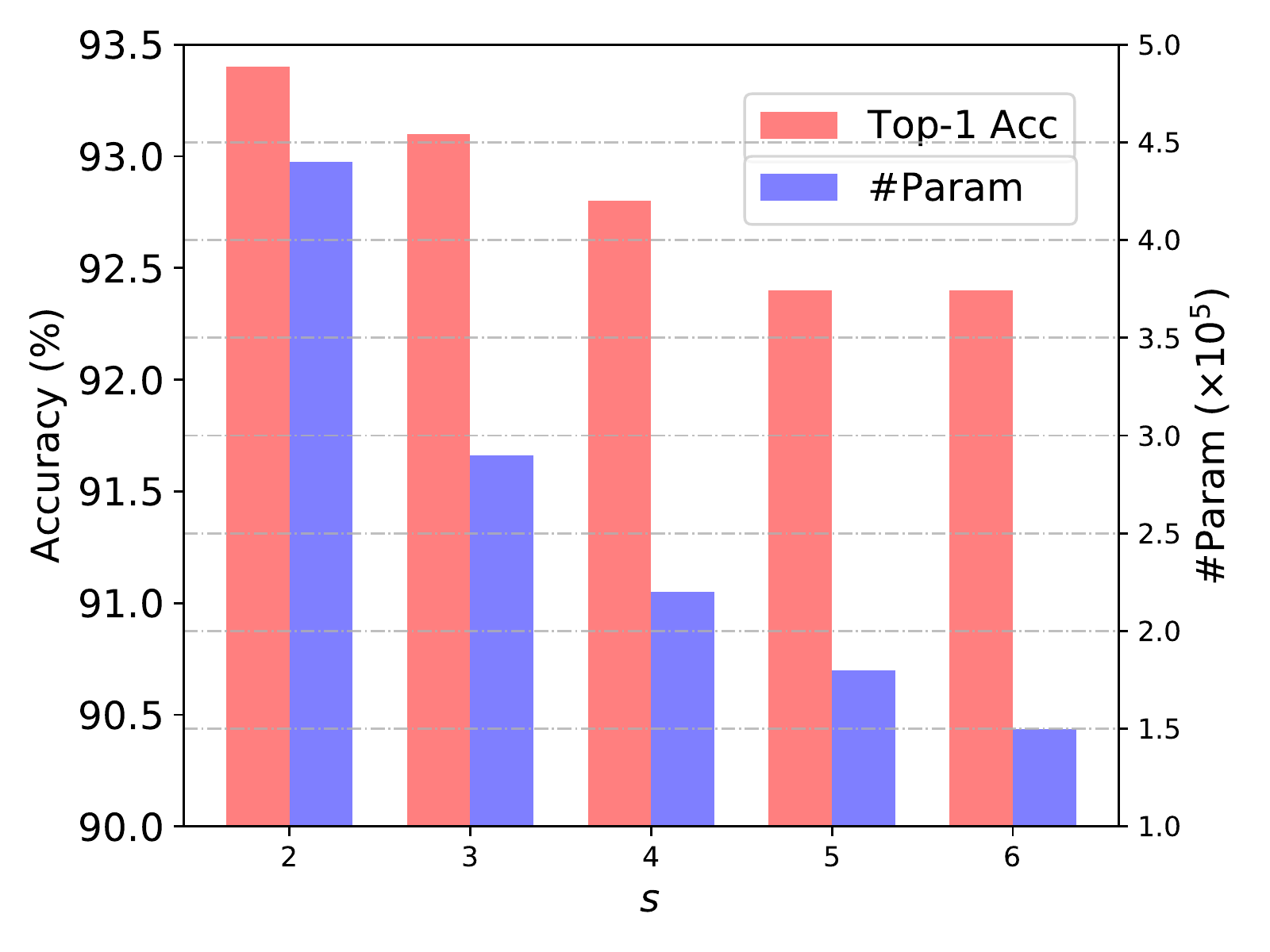}& \includegraphics[width=0.24\textwidth]{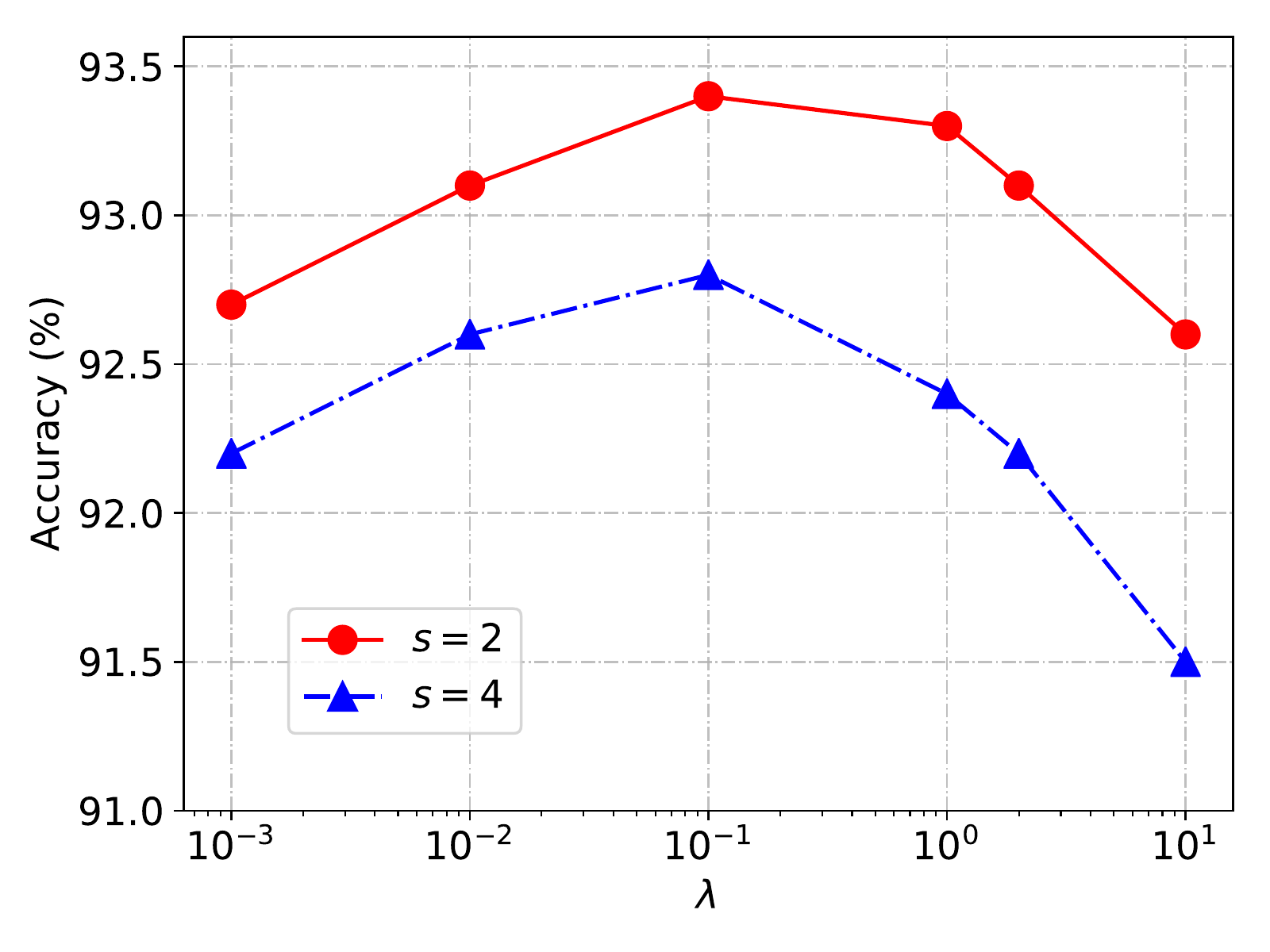}\\
		(a) Accuracy \emph{v.s.} $s$ & (b) Accuracy \emph{v.s.} $\lambda$ \\
	\end{tabular}
	\centering
	\caption{The performance of the proposed L-Versatile \emph{v.s.} $s$ and $\lambda$.} 
	\label{fig:ablation}
\end{figure}

\noindent\textbf{Impact of Hyper-parameters:}
We next investigate the influence of the compression ratio on the model performance on the CIFAR-10 dataset. ResNet-56 based separate L-Versatile filters is adopted here. In practice, $s$ is an important hyper-parameter directly controlling the compression ratio, and in this experiment we set $s$ in the range of $\{2,3,4,5,6\}$. The relation between the number of parameters and the classification accuracy is plotted in Fig. \ref{fig:ablation}(a). With the increase of $s$, the number of model parameters decreases rapidly, while the accuracy starts to decrease slightly. In the case where $s=6$, the final model only needs $1.5\times10^5$ parameters (compression ratio $\sim5.7\times$) with a $92.4\%$ classification accuracy. Thus our method promises a larger compression ratio meanwhile maintains a higher performance. The value of $s$ can be chosen according to the demand or restrictions of devices.

The orthogonal regularization is used for encouraging diversity of the masks. $\lambda$ is the hyper-parameter for the trade-off between orthogonal regularization and recognition loss. We set $s=\{5,10,20\}$ and tune $\lambda$ from $0.001$ to $10$. The results are reported in Fig.\ref{fig:ablation}(b). For different $s$, the experiment results are consistent with the above observation that smaller $s$ basically achieves better accuracy. The model obtains the best performance around $\lambda=0.1$ (in the following experiments for large-scale datasets, we set $\lambda=0.1$). When $\lambda$ is less than $0.1$, the accuracy decreases gradually; and as $\lambda$ increases from $0.1$, the accuracy stays steady at first and then decreases when $\lambda$ becomes too large to disturb the learning of the recognition loss. The model can obtain a satisfactory performance under a relatively wide range of $\lambda$, so we can choose an appropriate hyper-parameter easily in practice.

\subsection{Efficient Visual Classification}
We next evaluate the proposed method on the large scale image dataset, namely ImageNet ILSVRC 2012 dataset~\cite{ImageNet}, which contains over 1.2M training images and 50k validation images.

\begin{table}[ht]
	\begin{center}
		\renewcommand{\arraystretch}{1.05} 
		\caption{Statistics for versatile filters on ImageNet.}
		\label{Tab:AlexNet}
		\setlength{\tabcolsep}{1.0mm}
		\begin{tabular}{l||c|c|c|c|c}
			\hline
			\multirow{2}*{AlexNet} & \#Param & Mem & \#MUL & \multirow{2}*{Top1err} & \multirow{2}*{Top5err}\\
			& ($\times10^7$) & (MB) & ($\times10^8$) & & \\
			
			\hline
			Vanilla~\cite{alexnet} &  $6.1$ & 233 &  $7.2$ & $42.9\%$ & $19.8\%$\\
			S-Versatile & $3.5$ & 132 & $3.2$ & $42.1\%$ & $19.5\%$ \\
			S+C-Versatile & $1.9$ & 74 & $1.7$ & $44.1\%$ & $20.7\%$ \\
			Shared L-Versatile ($s$=4)  & $1.8$ & 70& $2.0$ & $43.7\%$ & $20.5\%$ \\
			Separate L-Versatile ($s$=4)  & $2.0$ & 76 & $2.0$ & $43.1\%$ & $19.9\%$ \\
			\hline
			\hline		
			\multirow{2}*{ResNet-50} & \#Param& Mem & \#MUL & \multirow{2}*{Top1err} & \multirow{2}*{Top5err}\\
			& ($\times10^7$) & (MB) & ($\times10^9$) & & \\
			\hline
			Vanilla~\cite{he2016deep} & $2.6$ & 97& $4.1$ & $24.7\%$ & $7.8\%$ \\
			S-Versatile & $1.9$ & 76 & $3.0$ & $ 24.5\%$ & $7.6\%$\\
			S+C-Versatile & $1.1$ & 42 & $1.5$& $25.5\%$ & $8.2\%$ \\
			Shared L-Versatile ($s$=4) & $0.8$ & 30 & $1.1$ & $26.8\%$ & $8.9\%$ \\
			Separate L-Versatile ($s$=4) & $0.9$ & 33 & $1.1$ & $25.5\%$ & $8.0\%$ \\
			Separate L-Versatile ($s$=32) & $0.36$ & 14 & $0.26$ & $26.5\%$ & $8.7\%$ \\
			\hline
		\end{tabular}
	\end{center}
\end{table}

\subsubsection{Large Neural Networks}
Two widely-used baseline architectures, AlexNet~\cite{alexnet} and ResNet-50~\cite{he2016deep}, were selected for conducting the following experiments. Note that, all training settings such as weight decay and learning rate used the same settings to ensure fair comparisons, \ie, SGD optimizer is used, \#epochs is 100, weight decay is 1e-4, learning rate starts from 0.1 and decays at 30-th, 60-th and 90-th epoch, data augmentation follows that in~\cite{he2016deep}.

\noindent\textbf{AlexNet:} AlexNet is one of the most classical deep CNN models for large scale visual recognition, which has over 230$\emph{MB}$ parameters and a $80.2\%$ accuracy on the ImageNet dataset with 1000 different categories. This network has 8 convolutional layers, sizes of convolution filters in the first six layers are larger than $1\times1$, \ie, $11\times11\times3\times96$, $5\times5\times48\times256$, $ 3\times3\times256\times384$, $3\times3\times192\times384$, $3\times3\times192\times256$, and $6\times6\times256\times4096$. 

Since sizes of convolution filters used in this network are much larger than that in other networks, resources required by this network can be significantly reduced by exploiting the proposed versatile convolution operation. For example, the parameter for the first convolutional layer is $s_1 = \lceil11/2\rceil = 6$, thus the number of parameters in this layer with versatile convolution filters is only $11\times11\times3\times16$. In this manner, we established a new network (S-Versatile in Table~\ref{Tab:AlexNet}) and reduced the number of filters in each convolutional layer according to its versatile parameter $s$. Specifically, sizes of convolution filters in its first six convolutional layers are $11\times11\times3\times16$, $5\times5\times48\times86$, $ 3\times3\times258\times192$, $3\times3\times192\times192$, $3\times3\times192\times128$, and $6\times6\times256\times1366$, respectively. After training the network on the ImageNet dataset, S-Versatile using Eq.~\ref{Fcn:MSConv} obtained a $19.5\%$ top5-err and a $42.1\%$ top-1 error rate, which are better than those of the baseline model. The memory usage of filters was reduced by a factor of $1.76\times$, and the number of multiplications in S-Versatile is $2.25\times$ less than that in the baseline model.

Furthermore, we applied the channel versatile filters (Eq.~\ref{Fcn:Channel}) on the S-Versatile AlexNet model with $\hat{c}-c=1$, and $g = 1$, namely, S+C-Versatile AlexNet. In this manner, the number of convolution layer in each layer will be reduced by a factor of $\frac{1}{2}$. As a result, this network achieved a $20.7\%$ top-5 error, which is slightly higher than that of the baseline model. But, the memory usage of the entire network is only $73.6$\emph{MB}, which is only about $30\%$ to that of the baseline model.

To evaluate the performance of L-Versatile on the large-scale dataset, we embed all the convolutional layers with L-versatile filters, except the last fully connected layer for classification (all the fully connected layers except the last one in AlexNet are regarded as convolutional layers). L-Versatile with shared masks forms Shared L-Versatile AlexNet, while L-Versatile with separate masks forms Separate L-Versatile AlexNet model in Table~\ref{Tab:AlexNet}. After training on the ImageNet dataset, the L-Versatile compresses AlexNet by $3-4\times$ and maintain the accuracy well. Similar to the observations in CIFAR-10 experiments, Separate L-Versatile performs better than Shared L-Versatile. Separate L-Versatile obtains $43.1\%$ top-1 error and $19.9\%$ top-5 error, which is much better than the S+C-Versatile model using spatial and channel versatile filters, indicating the advantages of learnable masks. 

\begin{table}[htb]
	\centering
	\renewcommand{\arraystretch}{1.1} 
	\caption{Comparison of the proposed versatile filters and other state-of-the-art model compression methods on ImageNet.}\label{tab:imagenet-sota}
	\setlength{\tabcolsep}{1.4mm}
	\begin{tabular}{l||c|c|c|c|c}
		\hline
		{Method} &  \#Param& \#ADD & \#MUL & \multirow{2}*{Top1err}  & \multirow{2}*{Top5err}\\
		& ($\times10^7$) & ($\times10^9$) & ($\times10^9$) & & \\
		\hline\hline
		Vanilla ResNet-50~\cite{he2016deep} &  $2.56$ & $4.1$ & $4.1$ & $24.7\%$ & $7.8\%$ \\
		Winograd~\cite{winograd} & $2.56$ & $3.1$ & $3.1$ & $24.7\%$ & $7.8\%$ \\
		NISP~\cite{nisp} & $1.44$ & $2.3$ & $2.3$ & - & $9.2\%$ \\
		SSS~\cite{sss} & - & $2.8$ & $2.8$ & $25.8\%$ & $8.2\%$ \\
		ThiNet-Conv \cite{Thinet} & - & $1.2$ & $1.2$ & $30.6\%$ & - \\
		ShiftResNet \cite{Wu_2018_CVPR} & $0.60$ & - & - & $29.4\%$ & $10.1\%$ \\
		Taylor-FO-BN~\cite{molchanov2019importance} & $0.79$ & $1.3$ & $1.3$ & $28.3\%$ & - \\
		GAL-1-joint~\cite{gal} & $1.02$ & $1.1$ & $1.1$ & $30.7\%$ & $10.9\%$ \\
		Slimmable 0.5$\times$ \cite{slimmable} & $0.69$ & $1.1$ & $1.1$ & $27.9\%$ & - \\
		MetaPruning~\cite{metapruning} & - & $1.0$ & $1.0$ & $26.6\%$ & - \\
		HRank~\cite{hrank} & $0.87$ & $1.6$ & $1.6$ & $28.0\%$ & $9.0\%$ \\		
		\hline
		Sep. L-Versatile ($s$=4) & $0.87$ & $2.1$ & $1.1$ & $25.5\%$ & $8.0\%$ \\
		\hline
	\end{tabular}
\end{table}

\begin{figure}[t]
	\centering
	\includegraphics[width=0.9\linewidth]{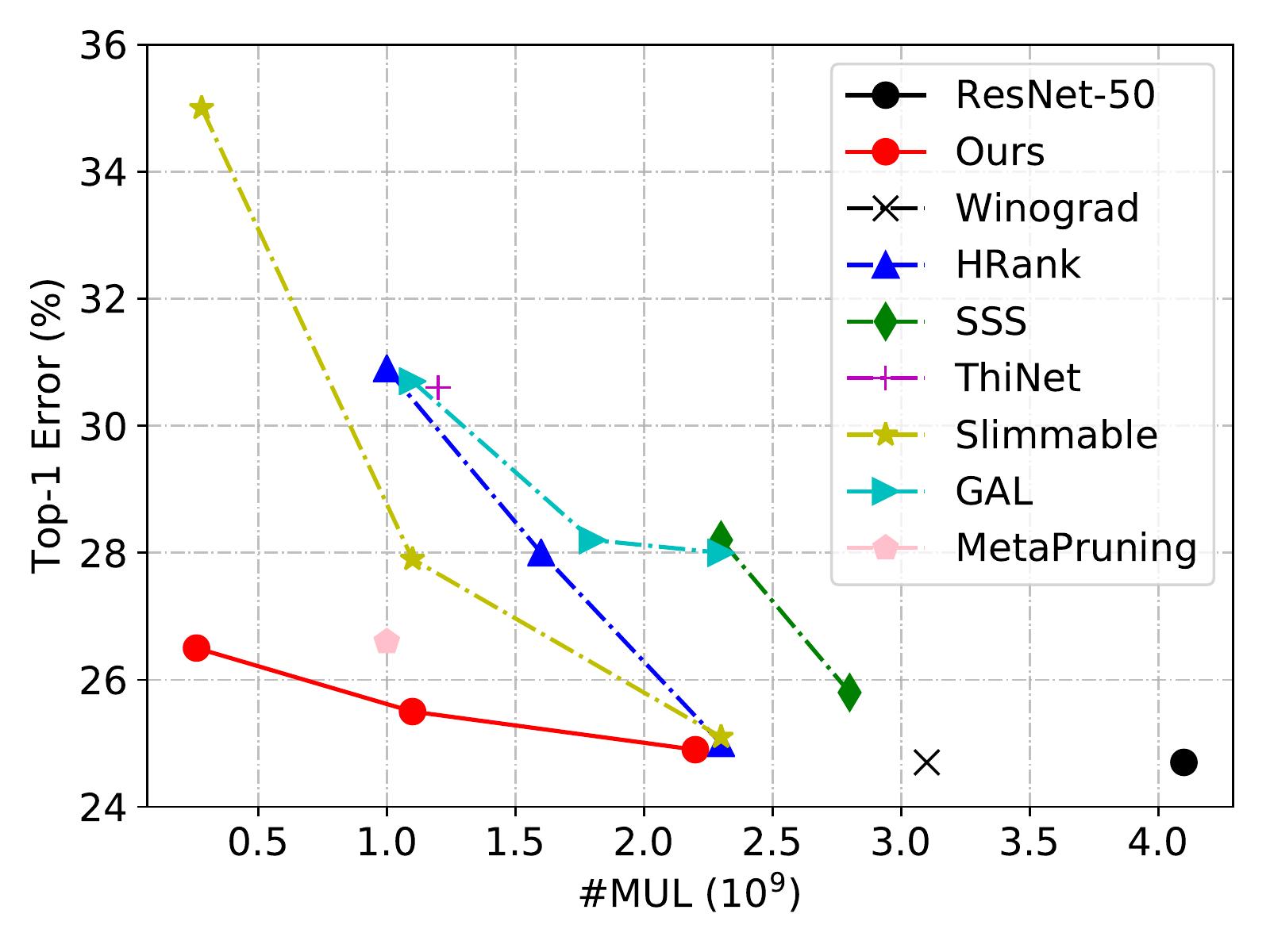} 
	\caption{ImageNet Top-1 Error vs. \#MUL.} 
	\label{fig:res50}
\end{figure}

\noindent\textbf{ResNet:} 
We reset the original convolutional layers in ResNets with the proposed versatile convolution filters. For instance, a convolutional layer of size $3\times3\times64\times128$ will be converted into a new layer of size $3\times3\times64\times32$ using the proposed versatile convolution filters. The performance of the original ResNet-50 and the network using versatile filters were detailed in Table~\ref{Tab:AlexNet}. There are still considerable filters in the ResNet whose sizes are larger than $1\times1$. Thus, its memory usage and FLOPs were reduced obviously by exploiting the proposed spatial versatile convolution filters. S-Versatile ResNet-50 with the same amount feature maps achieved a $7.6\%$ top-5 accuracy, which is slightly lower than that of the baseline models with only $75.6$\emph{MB} and $3.0\times10^9$ MULs.

In addition, S+C-Versatile ResNet-50 with the same volume of feature maps achieved a $8.2\%$ top-5 accuracy, which is slightly higher than that of the baseline models. Its memory usage is only about $41.7$\emph{MB}, which is only about $\frac{1}{2}$ to that of the original network. Therefore, our Versatile v2-ResNet-50, which is a more portable alternative to the original ResNet-50 model. Our versatile filters method provide a more flexible way for designing CNNs of high performance and portable architectures.

When setting $s=4$, Shared L-Versatile has an accuracy drop of about 1 point, while Separate L-Versatile has a better accuracy, achieving an $8.0\%$ top-5 error which is comparable with that of the baseline model. It is surprising to see that when we further increase $s$ to 32 in all the convolutional layers in Separate L-Versatile, the top-5 accuracy only decreases $0.9\%$ from that of the original model. In addition, our Separate L-Versatile model can reduce the number of multiplications by about $32\times$ as described in Section \ref{sec:mul}.

We also compare the proposed versatile filters with other state-of-the-art model compression methods, including filter level pruning~\cite{Thinet,sss,metapruning}, and new convolution operation~\cite{Wu_2018_CVPR}. The vanilla ResNet-50 trained on ImageNet is adopted as the baseline model. From the results in Table~\ref{tab:imagenet-sota} and Fig.~\ref{fig:res50}, we can see that with comparable or less parameters and computational cost, the proposed versatile filters method obtains $25.5\%$ top-5 error and $8.0\%$ top-1 error, which are better than the competitors significantly. Winograd's minimal filtering algorithms provide an effective acceleration of state-of-the-art convolutional neural networks with small 3$\times$3 filters \cite{winograd}, \eg, VGGNet. But nowadays modern CNNs often involve massive 1$\times$1 convolutions, \eg, ResNet and EfficientNet, which cannot be well accelerated by Winograd, as shown in Table~\ref{tab:imagenet-sota}. 

\noindent\textbf{Inference Latency:}
In order to test the speed-up effect of the multiplication reduction, we implement the separate learnable versatile filters using NCNN~\cite{ncnn}, and test it on an Intel i7-6950X CPU and Huawei Kirin 990 ARM CPU with single-thread mode. From Table~\ref{tab:latency}, we can see that Separate L-Versatile reduces the multiplications by 4$\times$ and speed-up the inference by about 2.4-2.5$\times$. To maximize the speed-up effect of the multiplication reduction in versatile filters, more engineering efforts could be taken to further optimize the implementation of versatile filters on hardware, \eg,~FPGA and NPU.

\begin{table}[htb]
	\centering
	\renewcommand{\arraystretch}{1.1} 
	\caption{Comparison of inference latency of ResNet networks.}\label{tab:latency}
	\setlength{\tabcolsep}{8pt}
	\begin{tabular}{l||c|c|c}
		\hline
		{Method} & \#MUL & CPU (ms) & Top1err \\		
		\hline\hline
		Vanilla ResNet-50            & $4.1$B & $405.2$  & $24.7\%$  \\
		Taylor-FO-BN~\cite{molchanov2019importance} & $1.3$B & $164.4$ & $28.3\%$ \\
		Slimmable 0.5$\times$ \cite{slimmable} & $1.1$B & $135.9$  & $27.9\%$ \\
		MetaPruning~\cite{metapruning} & $1.0$B & $132.4$ & $26.6\%$ \\
		Separate L-Versatile ($s$=4) & $1.1$B & $162.2$ & $25.5\%$ \\		
		\hline
	\end{tabular}
\end{table}

\noindent\textbf{Combination with Quantization:}
We combine quantization with our method for further model compression. The widely-used post-training quantization in TensorFlow~\cite{tensorflow} is adopted here. BitOps is utilized to measure the computational complexity~\cite{wu2018mixed,guo2020single} which is calculated as $\#BitOps=Bit_w\cdot Bit_a\cdot Ops$ where $Bit_w$, $Bit_a$ are the bit-widths of weights and activations. From the ImageNet results in Table~\ref{Tab:quant}, our 8-bit Separate L-Versatile ($s$=4) model achieves lower error rate than the original quantized models with similar BitOps.

\begin{table}[ht]
	\begin{center}
		\renewcommand{\arraystretch}{1.05} 
		\caption{Combination of the proposed versatile filters and post-training quantization method.}
		\label{Tab:quant}
		\setlength{\tabcolsep}{3.0mm}
		\begin{tabular}{l||c|c|c}
			\hline
			\multirow{2}*{Method} & Mem & \#BitOPs & \multirow{2}*{Top1err} \\
			& (MB)  & ($\times10^{10}$) & \\
			\hline\hline
			FP32 ResNet-50~\cite{he2016deep} & 97 & float point & $24.7\%$  \\
			FP32 Separate L-Versatile ($s$=4) & 33 & float point & $25.5\%$  \\
			\hline
			8-bit ResNet-50~\cite{tensorflow} & 24 & $26.2$ & $24.8\%$  \\
			6-bit ResNet-50~\cite{tensorflow} & 18 & $14.8$ & $25.7\%$  \\
			5-bit ResNet-50~\cite{tensorflow} & 15 & $10.3$ & $30.1\%$ \\
			8-bit Separate L-Versatile ($s$=4) & 9 & $9.9$ & $25.7\%$ \\
			\hline
		\end{tabular}
	\end{center}
\end{table}

\subsubsection{Small Neural Networks}
\noindent\textbf{MobileNet:} 
Besides sophisticate CNNs such as AlexNet and ResNet-50 with heavy architectures, a variety of recent works attempt to design neural networks with portable architectures and comparable performance, \eg, MobileNet~\cite{mobilenet,mobilev2} and ShuffleNet~\cite{shufflenet,shufflev2}. Unlike AlexNet or ResNet, these portable models already have more compact structure and less redundancy, thus further compression on them is more challenging.

We train our models under the similar settings as MobileNetV2~\cite{mobilev2}. We use the RMSProp optimizer with both decay and momentum of 0.9. The learning rate starts from 0.048 and decays by 0.97 per 2.4 epochs. The weight decay is set as 1e-5. 8 GPUs are used with 128 images per chip. Data augmentation follows that in~\cite{inceptionv3}. Table~\ref{tab:portable} summarizes state-of-the-art CNN architectures, including their memory usages, FLOPs, and recognition results on the ILSVRC 2012 dataset. Obviously, these portable models have much smaller model size and fewer FLOPs than AlexNet or ResNet, but its classification accuracy is acceptable. We embed the proposed versatile filters on the state-of-the-art portable architecture, \ie,~MobileNet-v2. Since most of the weights and FLOPs are occupied by the $1\times1$ pointwise convolutions, the channel versatile filters and versatile filters with learnable masks are applicable here. By exploiting the proposed channel versatile convolution filters on the MobileNet-v2, we reduced more than $30\%$ weights of convolution filters and achieved a comparable accuracy, which is a more portable convolutional neural network. By exploiting the separate learnable masks on MobileNet-v2 with $s=4$, the performance of Separate L-Versatile has a small decrease but is still better than that of other portable models with a smaller model size (7.5 MB) and significantly fewer multiplication operations (93 M).

\begin{table}[htb]
	\centering
	\renewcommand{\arraystretch}{1.1} 
	\caption{Comparison with state-of-the-art portable CNNs on ImageNet.}\label{tab:portable}
	\setlength{\tabcolsep}{0.9mm}
	\begin{tabular}{l||c|c|c|c}
		\hline
		{Method} & \#Param & \#ADD & \#MUL & \multirow{2}*{Top1err} \\
		& ($\times10^6$) & ($\times10^8$) & ($\times10^8$) & \\
		
		\hline\hline
		ShuffleNet-v1 1.5$\times$ (g=3)~\cite{shufflenet}   & $3.4$& $2.9$   & $2.9$     & $31.0\%$   \\
		ShuffleNet-v1 1$\times$ (g=8)~\cite{shufflenet} & $2.5$ & $1.4$ & $1.4$ & $32.4\%$  \\
		ShuffleNet-v2 1$\times$~\cite{shufflev2} & $2.3$ & $1.5$ & $1.5$ & $30.6\%$  \\
		ChannelNet-v2~\cite{gao2018channelnets} & $2.7$ & $3.6$ & $3.6$ & $30.5\%$ \\
		AS-ResNet-w50~\cite{active-shift} & $2.0$ & $4.0$ & $4.0$ & $30.1\%$ \\
		MobileNet-v2 0.75$\times$~\cite{mobilev2} & $2.6$& $2.1$ & $2.1$ & $30.2\%$  \\
		MobileNet-v2 1.0$\times$~\cite{mobilev2} & $3.5$ & $3.0$ & $3.0$ & $28.2\%$ \\
		\hline
		C-Versatile \tiny{(MobileNet-v2 1.0$\times$)}  & $2.4$ & $1.6$ & $1.6$ & $29.5\%$ \\
		Separate L-Versatile \tiny{($s$=4, MobileNet-v2 1.0$\times$)} & $2.0$ & $1.6$ & $0.9$ & $29.7\%$ \\		
		\hline
	\end{tabular}
\end{table}

\begin{figure*}[t]
	\setlength{\tabcolsep}{1.5mm}
	\centering
	\begin{tabular}{ccc}
		\includegraphics[width=0.3\textwidth]{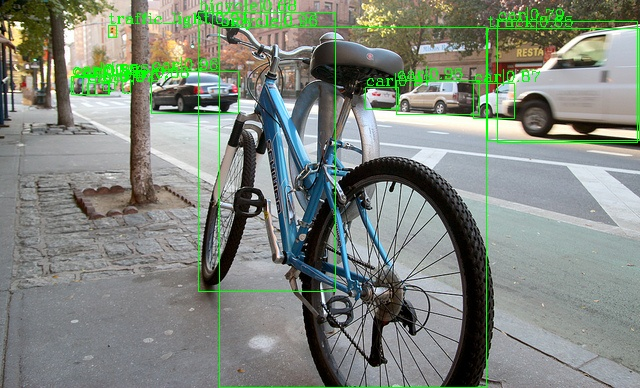}& \includegraphics[width=0.3\textwidth]{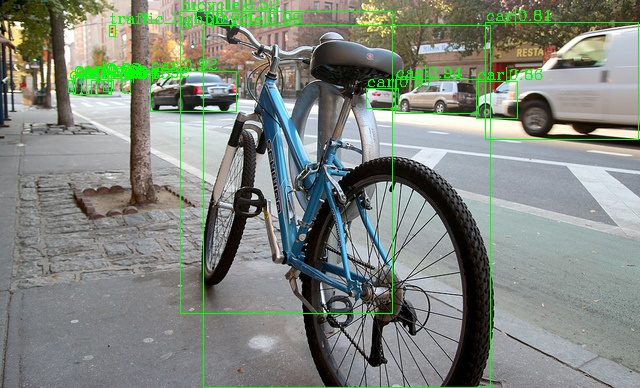}& \includegraphics[width=0.3\textwidth]{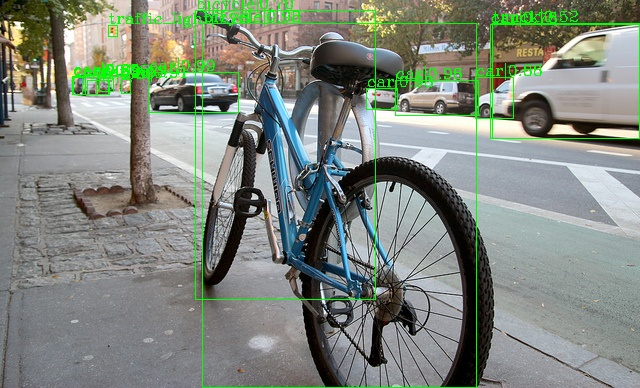}\\
		(a) Vanilla & (b) S-Versatile & (c) Separate L-Versatile\\
		\includegraphics[width=0.3\textwidth]{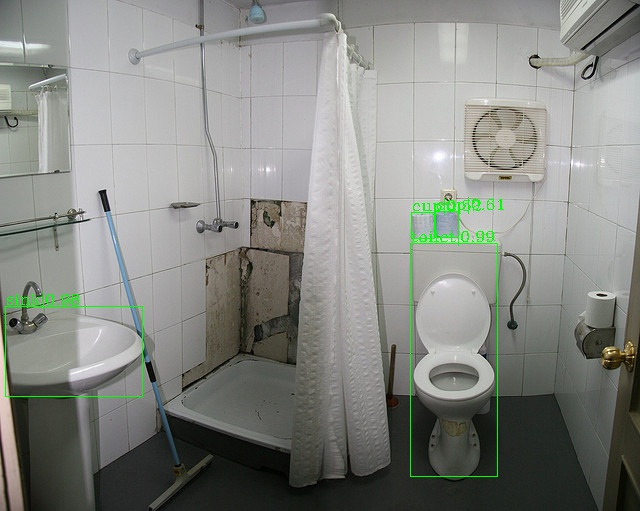}& \includegraphics[width=0.3\textwidth]{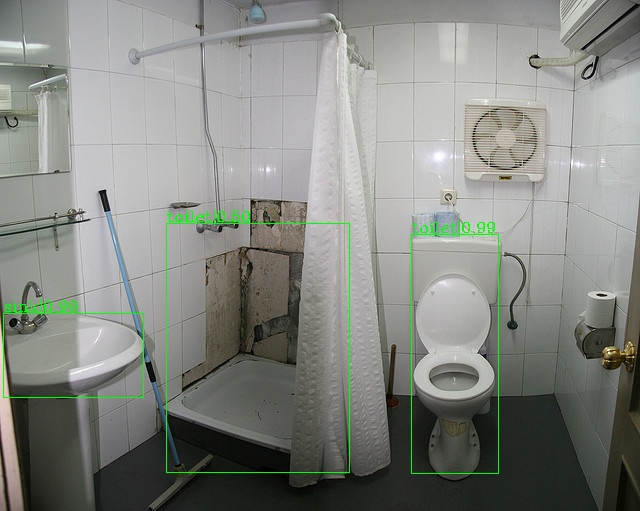}& \includegraphics[width=0.3\textwidth]{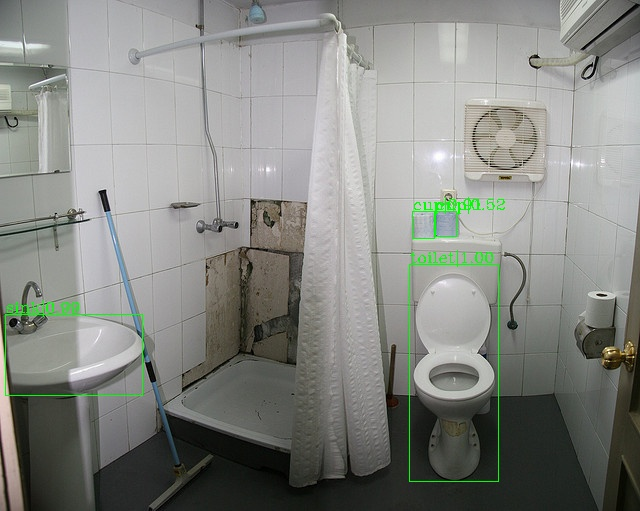}\\
		(d) Vanilla & (e) S-Versatile & (f) Separate L-Versatile\\
	\end{tabular}
	\caption{Example objection detection results on MS COCO (The figure is better viewed in color).} 
	\label{fig:detect-vis}
\end{figure*}

\noindent\textbf{Inference Latency:}
We also test the actual inference latency of small CNNs on Huawei Kirin 980 ARM CPU with single-thread mode. From Table~\ref{tab:mv2-latency}, our Separate L-Versatile can largely accelerate MobileNet-v2 and achieve the fastest inference with the lowest top-1 error rate.

\begin{table}[htb]
	\centering
	\renewcommand{\arraystretch}{1.1} 
	\caption{Comparison of inference latency of small networks.}\label{tab:mv2-latency}
	\setlength{\tabcolsep}{2pt}
	\begin{tabular}{l||c|c|c|c}
		\hline
		\multirow{2}*{Method} & \#ADD & \#MUL & Latency & \multirow{2}*{Top1err} \\
		& ($\times10^8$) & ($\times10^8$) & (ms) & \\	
		\hline\hline
		ShuffleNet-v1 1.5$\times$ (g=3)~\cite{shufflenet} & $2.9$ & $2.9$ & $76.8$  & $31.0\%$  \\
		ShuffleNet-v2 1$\times$~\cite{shufflev2} & $1.5$ & $1.5$ & $30.8$ & $30.6\%$ \\
		ChannelNet-v2~\cite{gao2018channelnets} & $3.6$ & $3.6$ & $58.6$ & $30.5\%$ \\
		MobileNet-v2 0.75$\times$~\cite{mobilev2} & $2.1$ & $2.1$ & $36.8$ & $30.2\%$ \\
		Separate L-Versatile \tiny{($s$=4, MobileNet-v2 1.0$\times$)} & $1.6$ & $0.9$ & $27.0$ & $29.7\%$ \\		
		\hline
	\end{tabular}
\end{table}

\noindent\textbf{EfficientNet:}
EfficientNet~\cite{efficientnet}  is the state-of-the-art network architecture discovered through the neural architecture search~\cite{nas1,mnasnet}. To evaluate the generalization of our versatile filters, we apply it on EfficientNet-B0 with 390M FLOPs. All the $1\times1$ point-wise convolutional layers in inverted residual blocks are equipped with our versatile filters. We train our models under the similar settings as EfficientNet~\cite{efficientnet}. We use the standard RMSProp optimizer with both decay and momentum of 0.9. The learning rate starts from 0.045 and decays by 0.98 per epoch. Data augmentation follows that in~\cite{inceptionv3}. Model EMA (exponential moving average) is also used.  8 GPUs are used with 128 images per chip. The results are shown in Table~\ref{tab:nas}. The channel versatile filters can reduce about more than $30\%$ weights with a slight accuracy drop. The versatile filters with separate learnable masks ($s = 2$) can improve the performance to $22.9\%$ top-1 error which outperforms those state-of-the-art NAS methods by a large margin. 

\begin{table}[htb]
	\centering
	\renewcommand{\arraystretch}{1.1} 
	\caption{Comparison with NAS searched efficient neural networks on ImageNet.}\label{tab:nas}
	\setlength{\tabcolsep}{1.3mm}
	\begin{tabular}{l||c|c|c|c}
		\hline
		{Method} & \#Param & \#ADD & \#MUL & \multirow{2}*{Top1err} \\
		& ($\times10^6$) & ($\times10^8$) & ($\times10^8$) & \\		
		\hline\hline
		Vanilla EfficientNet-B0~\cite{efficientnet} & $5.3$ & $3.9$ & $3.9$ & $22.7\%$ \\
		ProxylessNAS~\cite{proxylessnas} & $4.1$ & $3.2$ & $3.2$ & $25.4\%$ \\
		MnasNet-A1~\cite{mnasnet} & $3.9$ & $3.2$ & $3.1$ & $24.8\%$ \\
		FBNet~\cite{fbnet} & $4.5$ & $3.0$ & $3.0$ & $25.9\%$ \\
		FBNetV2-F4~\cite{fbnetv2} & $-$ & $2.4$ & $2.4$ & $24.0\%$ \\
		MobileNetV3 Large 1.0$\times$~\cite{mobilenetv3} & $5.4$ & $2.2$ & $2.2$ & $24.8\%$ \\
		MobileNetV3 Large 0.75$\times$~\cite{mobilenetv3} & $4.0$ & $1.6$ & $1.6$ & $26.7\%$ \\
		\hline
		C-Versatile \tiny{(EfficientNet-B0)}  & $3.4$ & $2.1$ & $2.1$ & $23.1\%$ \\
		Separate L-Versatile \tiny{($s$=2, EfficientNet-B0)} & $3.5$ & $2.1$ & $2.1$ & $22.9\%$ \\	
		Separate L-Versatile \tiny{($s$=4, EfficientNet-B0)} & $2.5$ & $2.1$ & $1.2$ & $23.2\%$ \\		
		\hline
	\end{tabular}
\end{table}

\subsection{Efficient Visual Detection}
To investigate the generalization ability of the proposed versatile filters, we further evaluate it on the MS COCO object detection task \cite{lin2014microsoft}. We adopt Faster-RCNN \cite{renNIPS15fasterrcnn} with the short edge size of 600 as the detection framework and use ResNet-50 as the backbone architecture. The models are trained for 12 epochs with initial learning rate 0.01 which is decayed at 8-th and 11-th epoch. The COCO train+val dataset excluding 5,000 minival images are used for training and the minival set is for testing. From the results in Table~\ref{tab:coco}, spatial versatile filters reduce the weights and multiplications significantly meanwhile maintaining the mAP. The L-Versatile can compress ResNet-50 by $3.5\times$ and obtain comparable performance.

We give some examples of object detection results on MS COCO for further comparison. It can be seen from Fig.\ref{fig:detect-vis} that the difference of results between the original Faster R-CNN and these with versatile filters is negligible small. Especially in the second row, the original Faster R-CNN model missed the ``toilet'', and the Separate L-Versatile model performs the same, while the S-Versatile model succeed to detect the ``toilet'' due to its multi-scale receptive field in one convolutional layer.

\begin{table}[htbp]
	\centering
	\renewcommand{\arraystretch}{1.1} 
	\caption{The object detection results on MS COCO. mAP is reported with COCO primary challenge metric (AP at IoU=0.50:0.05:0.95)} \label{tab:coco}
	\small\vspace{0.1cm}
	\begin{tabular}{l||c|c|c}
		\hline
		\multirow{2}{*}{Method} & {Backbone} & Backbone & \multirow{2}{*}{mAP} \\
		& \#Param & \#MUL &\\
		\hline\hline
		Vanilla Faster R-CNN \cite{renNIPS15fasterrcnn}  & $2.4\times 10^7$ & $5.0\times 10^9$ & $33.1\%$ \\
		S-Versatile  &  $1.7\times 10^7$  & $3.7\times 10^9$ & $33.0\%$ \\
		Separate L-Versatile ($s$=4)  &  $0.7\times 10^7$  & $1.3\times 10^9$ & $32.1\%$ \\
		\hline
	\end{tabular}
	\vspace{-0.3cm}
\end{table}

\subsection{Efficient Visual Super-Resolution}

\begin{figure*}[t]
	\begin{center}
		\setlength{\tabcolsep}{2pt}
		\begin{tabular}{ccccc}
			Input & Ground-truth & VDSR & S-Versatile & Separate L-Versatile\\
			\includegraphics[width=0.19\textwidth]{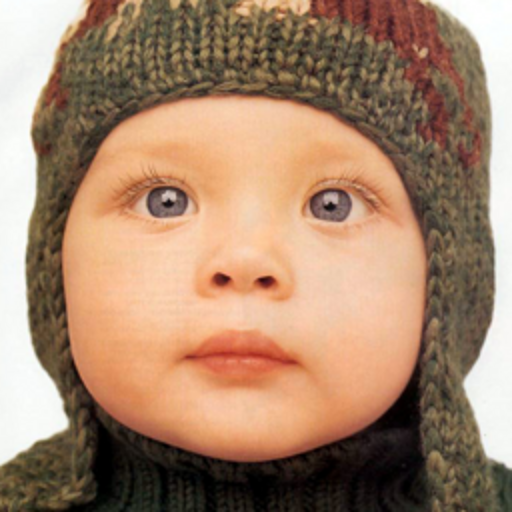}&
			\includegraphics[width=0.19\textwidth]{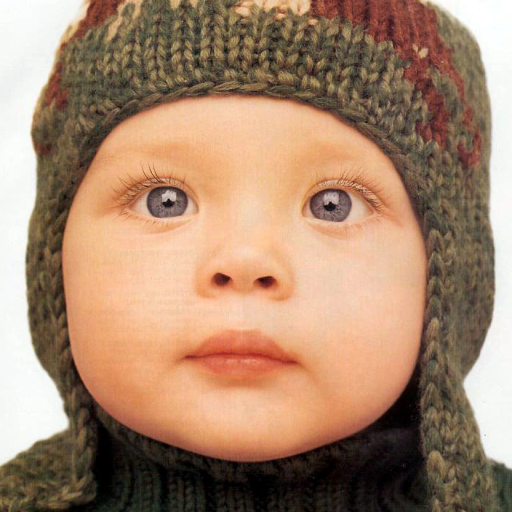}&
			\includegraphics[width=0.19\textwidth]{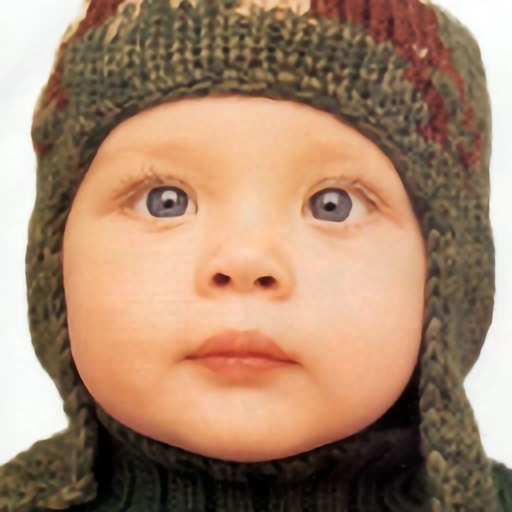}&
			\includegraphics[width=0.19\textwidth]{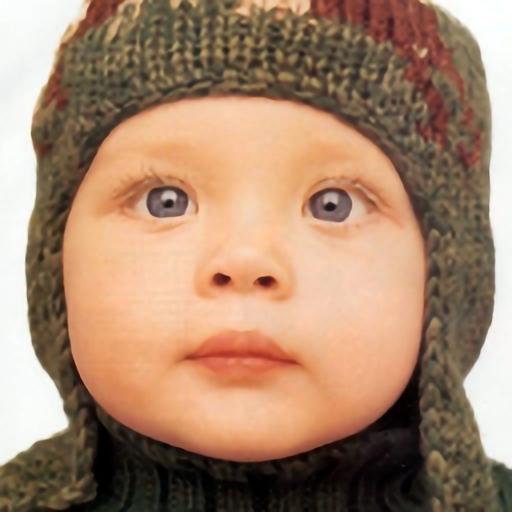}&
			\includegraphics[width=0.19\textwidth]{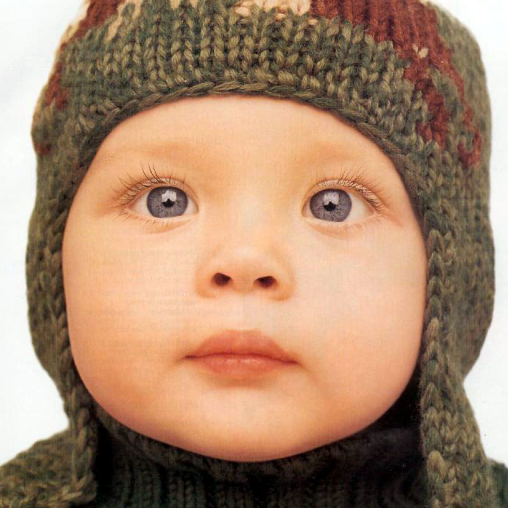}\\
			\emph{Baby} ($\times4$) &  & PSNR = $33.40$\emph{dB}& PSNR = $33.41$\emph{dB} & PSNR = $33.35$\emph{dB}\\
			\includegraphics[width=0.19\textwidth]{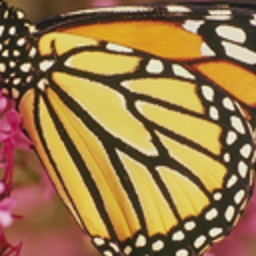}&
			\includegraphics[width=0.19\textwidth]{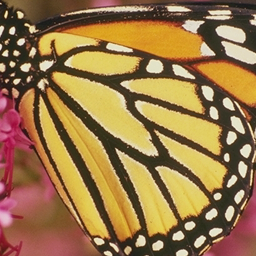}&
			\includegraphics[width=0.19\textwidth]{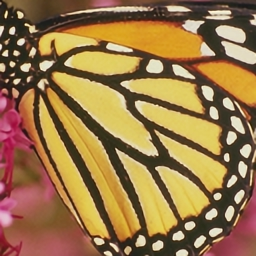}&
			\includegraphics[width=0.19\textwidth]{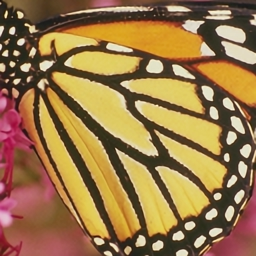}&
			\includegraphics[width=0.19\textwidth]{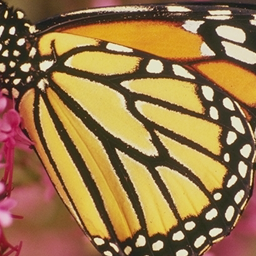}\\
			\emph{Butterfly} ($\times2$) & & PSNR = $34.45$\emph{dB} & PSNR = $34.51$\emph{dB}& PSNR = $34.42$\emph{dB}\\  	
		\end{tabular}
		\caption{Image super-resolution results of the baseline VDSR model and the proposed versatile filters, where the top line are results of the \emph{Baby} ($\times4$) image, and the bottom line are results of the \emph{Butterfly} ($\times2$) image.}
		\label{Fig:sr}
	\end{center}
	\vspace{-1em}
\end{figure*}

We applied the proposed versatile filters on the single image super-resolution problem. The image super-resolution task receives a low-resolution image and then outputs its high-resolution estimation. 
We selected VDSR (Very Deep CNN for Image Super-resolution~\cite{VDSR}) as the baseline model for conducting the image super-resolution task. The baseline model contains 22 convolutional layers with a number of $3\times3$ convolution filters, which was trained on a benchmark dataset consists of 291 images. Each image in this dataset is first divided into several patches and then augmented with some commonly used strategies (\ie,~rotation and flip) to form the training set. Although the VDSR model utilizes a relatively small dataset, but shows better performance than that of SRCNN~\cite{SRCNN} trained on the ILSVRC dataset due to it contains more convolutional layers. 

Similar to experiments in the main body, a new model using the proposed spatial versatile filters and another model using the proposed channel versatile filters were established, respectively. Then, the baseline VDSR model and the two new models were trained on the dataset using the same setting (\eg,~learning rate, number of epochs) used in~\cite{VDSR}, respectively. Images in the dataset were downscaled by $2\times$ and $4\times$ in order to train models for processing images with different resolutions. Detailed results are shown in Table~\ref{Tab:sr}. PSNR values on Set5~\cite{set5} were calculated by comparing output images and ground-truth high-resolution images.

\begin{table}[h]
	\begin{center}
		\renewcommand{\arraystretch}{1.1} 
		\caption{Statistics for versatile filters on VDSR.}
		\label{Tab:sr}
		\begin{tabular}{l|c|c|c}
			\hline
			Model & Memory &  PSNR ($\times2$) & PSNR ($\times4$) \\
			\hline
			\hline
			Vanilla VDSR~\cite{VDSR} & $2.82$\emph{MB} & $37.53$\emph{dB} &  $31.35$ \emph{dB}\\
			S-Versatile & $1.41$\emph{MB} &  $37.55 $\emph{dB}&  $31.28$ \emph{dB}\\
			Separate L-Versatile ($s$=4) & $0.73$\emph{MB} & $37.50$\emph{dB}&  $31.24$ \emph{dB}\\
			\hline			
		\end{tabular}
	\end{center}
\end{table}

It can be found in Table~\ref{Tab:sr} that, memory usage and computational complexity of networks using the proposed versatile convolution operation have been reduced significantly, and PSNR values of S-Versatile-VDSR with the same amount of feature maps are higher than those of the baseline model, while the memory usage of this model is only about $1.41$\emph{MB}. In addition, the memory usage of the Separate L-Versatile model is only about $0.73$\emph{MB} with the similar performance to that of the original model.

Fig.~\ref{Fig:sr} illustrates some visualization results of the baseline model and the proposed method. Results generated by networks using the proposed versatile convolution filters are better than those of the baseline model since we can provide the same amount feature maps with multi-scale information, which is able to make the estimation smooth in every scale.

\section{Conclusion}
Exploring convolutional neural networks with low memory usage and computational complexity is very essential so that these models can be used on mobile devices. In fact, the main waste in a general neural network is that a convolution filter with massive parameters can only produce one feature for a given data. In order to make full use of convolution filters, this paper proposes versatile convolution filters from spatial and channel perspectives. Thus, we can use fewer parameters to generate the same amount of useful features with a lower computational complexity at the same time. Experiments conducted on benchmark image datasets and models show that the proposed method can not only reduce the requirement of storage and computational resources, but also enhance the performance of CNNs, which is very effective for establishing portable CNNs with high accuracies. In addition, the proposed method can be easily implemented using the existing convolution component, we will further embed it into other applications such as image segmentation and image generation.

\section*{Acknowledge}
This work was supported by NSFC (61632003, 62072449) and Macau FDCT Grant  (Grant No. 0018/2019/AKP, SKL-IOTSC-2021-2023). This work was also supported in part by the Australian Research Council under Projects DE180101438 and DP210101859.

\ifCLASSOPTIONcaptionsoff
  \newpage
\fi

\renewcommand\refname{References}

{\small
\bibliographystyle{ieee}
\bibliography{ref}
}

\end{document}